%% file: submission-latex-2026.tex
\documentclass[letterpaper]{article} 
\usepackage{aaai2026}  
\usepackage{times}  
\usepackage{helvet}  
\usepackage{courier}  
\usepackage[hyphens]{url}  
\usepackage{graphicx} 
\usepackage{natbib}  
\usepackage{caption} 
\usepackage{algorithm}
\usepackage{algorithmic}

\usepackage{newfloat}
\usepackage{listings}
\DeclareCaptionStyle{ruled}{labelfont=normalfont,labelsep=colon,strut=off} 
\floatstyle{ruled}
\newfloat{listing}{tb}{lst}{}
\floatname{listing}{Listing}
\usepackage{nicematrix}
\usepackage{tikz}
\usepackage{inconsolata}
\usepackage{dsfont}
\usepackage{stmaryrd}
\usepackage{array,multirow}
\usepackage{rotating}
\usepackage[utf8]{inputenc}
\usepackage{csquotes}
\usepackage{tabularx}
\usepackage{kotex}
\usepackage{makecell}
\usepackage{booktabs}
\usepackage{pifont}
\usepackage{array}
\usepackage{soul}
\usepackage{xcolor,colortbl}

\definecolor{right}{RGB}{0,128,96}
\definecolor{wrong}{RGB}{192,0,32}
\definecolor{green}{RGB}{59, 125, 35}
\definecolor{skyblue}{RGB}{33, 95, 154}
\definecolor{orange}{RGB}{255, 102, 0}

\newcommand{\Right}[1]{\textcolor{right}{#1}}
\newcommand{\Wrong}[1]{\textcolor{wrong}{#1}}
\newcommand{\cmark}{\Right{\ding{51}}}
\newcommand{\xmark}{\Wrong{\ding{55}}}

\title{MAVIS: A Benchmark for Multimodal Source Attribution in\\Long-form Visual Question Answering}
\author{
    Seokwon Song$^{1}$, 
    Minsu Park$^{1}$, 
    Gunhee Kim$^{1}$\footnote{Corresponding author.}
}
\affiliations{
    \textsuperscript{\rm 1}Seoul National University\\

    \{seokwon.song, minsu.park\}@vision.snu.ac.kr, gunhee@snu.ac.kr
}

\usepackage{bibentry}

\begin{document}

\maketitle

\begin{abstract}

Source attribution aims to enhance the reliability of AI-generated answers by including references for each statement, helping users validate the provided answers. However, existing work has primarily focused on \textit{text-only scenario} and largely overlooked the role of \textit{multimodality}.
We introduce MAVIS, the first benchmark designed to evaluate multimodal source attribution systems that understand user intent behind visual questions, retrieve multimodal evidence, and generate long-form answers with citations.
Our dataset comprises 157K visual QA instances, where each answer is annotated with fact-level citations referring to multimodal documents.
We develop fine-grained automatic metrics along three dimensions of informativeness, groundedness, and fluency, and demonstrate their strong correlation with human judgments. Our key findings are threefold: (1) LVLMs with multimodal RAG generate more informative and fluent answers than unimodal RAG, but they exhibit weaker groundedness for image documents than for text documents, a gap amplified in multimodal settings. (2) Given the same multimodal documents, there is a trade-off between informativeness and groundedness across different prompting methods. (3) Our proposed method highlights mitigating contextual bias in interpreting image documents as a crucial direction for future research.
\end{abstract}

\begin{links}
     \link{Code}{https://github.com/seokwon99/MAVIS}
\end{links}

\section{Introduction}
\begin{figure}[t!]
\centering
\includegraphics[width=\linewidth]{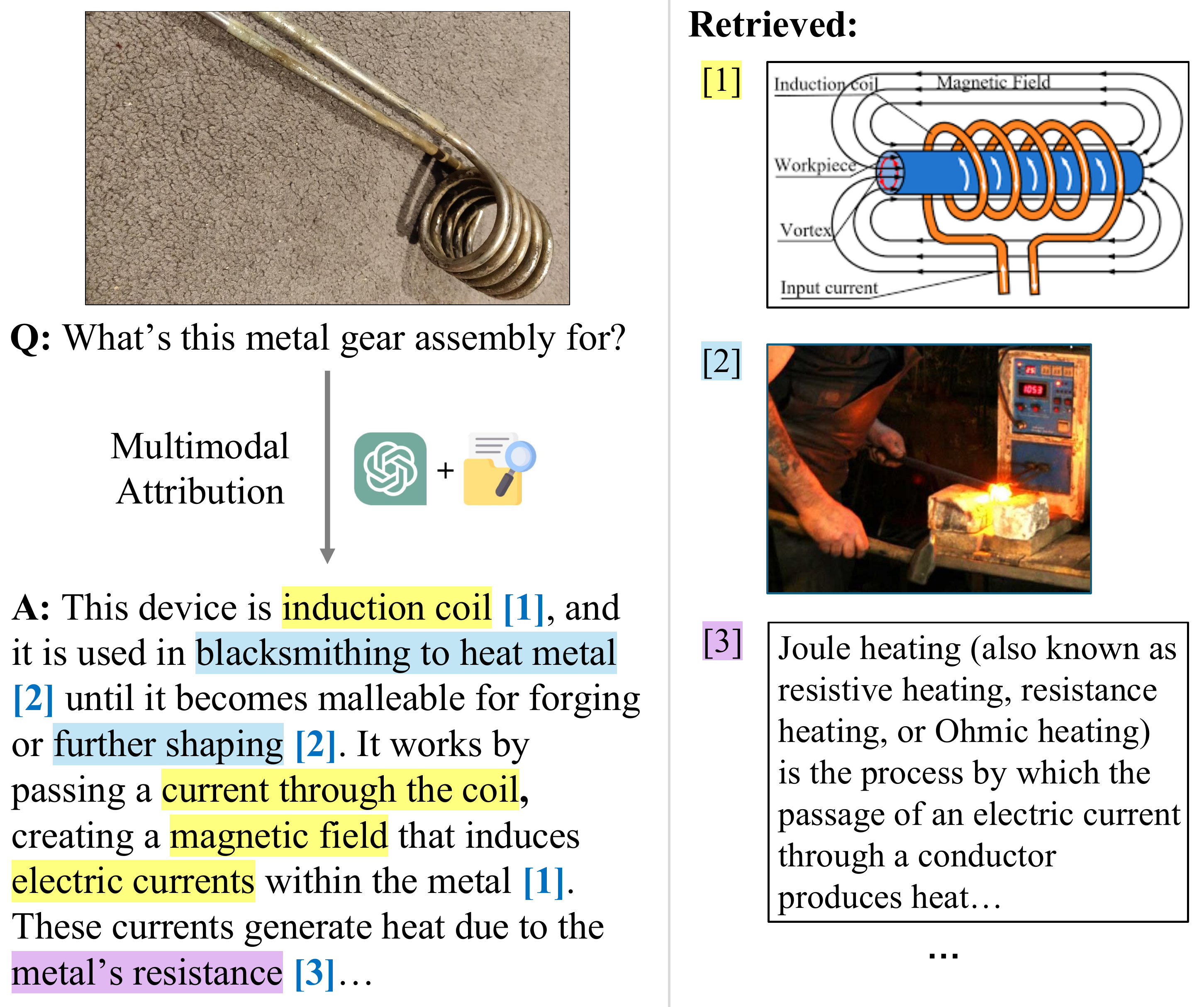}
\caption{Example of the system's response in our MAVIS benchmark. Given a user question paired with an input image, the system must generate a long-form answer supported by sentence-level citations that reference multimodal documents. Highlighted texts can be verified using the corresponding colored documents.}
\label{fig:intro_figure}
\end{figure}

\input{Tables/data_compare}

Since the advent of open-domain generative models, detecting and mitigating potentially inaccurate or fabricated information has become a critical challenge~\cite{ye2023cognitive, zhang2023siren}. This so-called hallucination problem arises from the models’ ability to generate highly fluent, human-like responses, making inaccuracies difficult to detect~\cite{huang2025survey}. This issue is particularly problematic in long-form generation, where the generated content frequently includes numerous pieces of information that are a mixture of true and false~\cite{min2023factscore}.

These challenges suggest a growing need that not only generates factually accurate answers but also supports them with verifiable evidence. One such approach includes source attribution, in which systems enhance the verifiability of their long-form answers by including citations for each statement~\cite{bohnet2022attributed, gao2023enabling}. Existing works in this area focus on the \textit{text-only} setting, where systems interpret user queries expressed in natural language and retrieve evidence solely from textual sources.

However, text-only attribution may be insufficient, as many real-world scenarios require multimodal understanding.
First, images are inherently \textit{compact yet rich}, making them suitable for conveying dense and detailed information intuitively. For instance, as shown in document [1] of Figure~\ref{fig:intro_figure}, the image simultaneously represents multiple layers of information—such as ``the structure of an induction coil'', ``the generation of a magnetic field by current flowing through the coil'', and ``the magnetic field inducing electric currents in metal''—in a compact and intuitive manner.
Second, user-provided images play a critical role in understanding user intent and enabling precise responses. It is often difficult to grasp the intent of a question based solely on a natural language query—for example, ``what's this metal gear assembly for?''—without  considering the accompanying image.

To bridge the gap between existing attribution benchmarks and real-world scenarios, we introduce MAVIS (\textbf{M}ultimodal \textbf{A}ttribution for \textbf{Vis}ual Question Answering), a benchmark designed to evaluate models on their ability to: (1) comprehend questions involving visual inputs, (2) generate effective search queries to retrieve multimodal documents, and (3) provide long-form answers with appropriate citations. We automatically construct a dataset of 157K instances, each containing a visual question and a ground-truth answer citing multimodal documents at the sentence level, and manually annotate 1K instances for evaluation. As shown in Table~\ref{tab:dataset_comparison}, our dataset uniquely includes both visual questions and multimodal supporting documents, and provides fact-level citation.


To ensure fine-grained evaluation of long-form responses and citation quality, we employ three metrics. (1) \textit{Informativeness} measures how thoroughly the answer covers necessary information without unnecessary content; (2) \textit{Groundedness} assesses how well the answer is supported by the citations; and (3) \textit{Fluency} evaluates how fluent and coherent the output is. Our human evaluation results show that this automatic evaluation highly correlates  with human judgments, making it a reliable evaluation method.

Through extensive experiments, we also find multiple intriguing observations. 
\begin{enumerate}
    \item Multimodal RAG generates more informative and fluent answers than unimodal RAG (i.e., text-only or image-only). However, LVLMs primarily rely on text when generating answers, making them less attentive to image-based documents.
    \item Given identical multimodal documents, there is a trade-off between informativeness and groundedness across different prompting methods, indicating the challenge of improving both simultaneously.
    \item LVLMs fabricate information more frequently from image documents than from text documents. We introduce a knowledge extraction step before the final answer generation, effectively addressing this issue by mitigating contextual bias.
\end{enumerate}

\begin{figure*}[ht!]
\centering
\includegraphics[width=\linewidth]{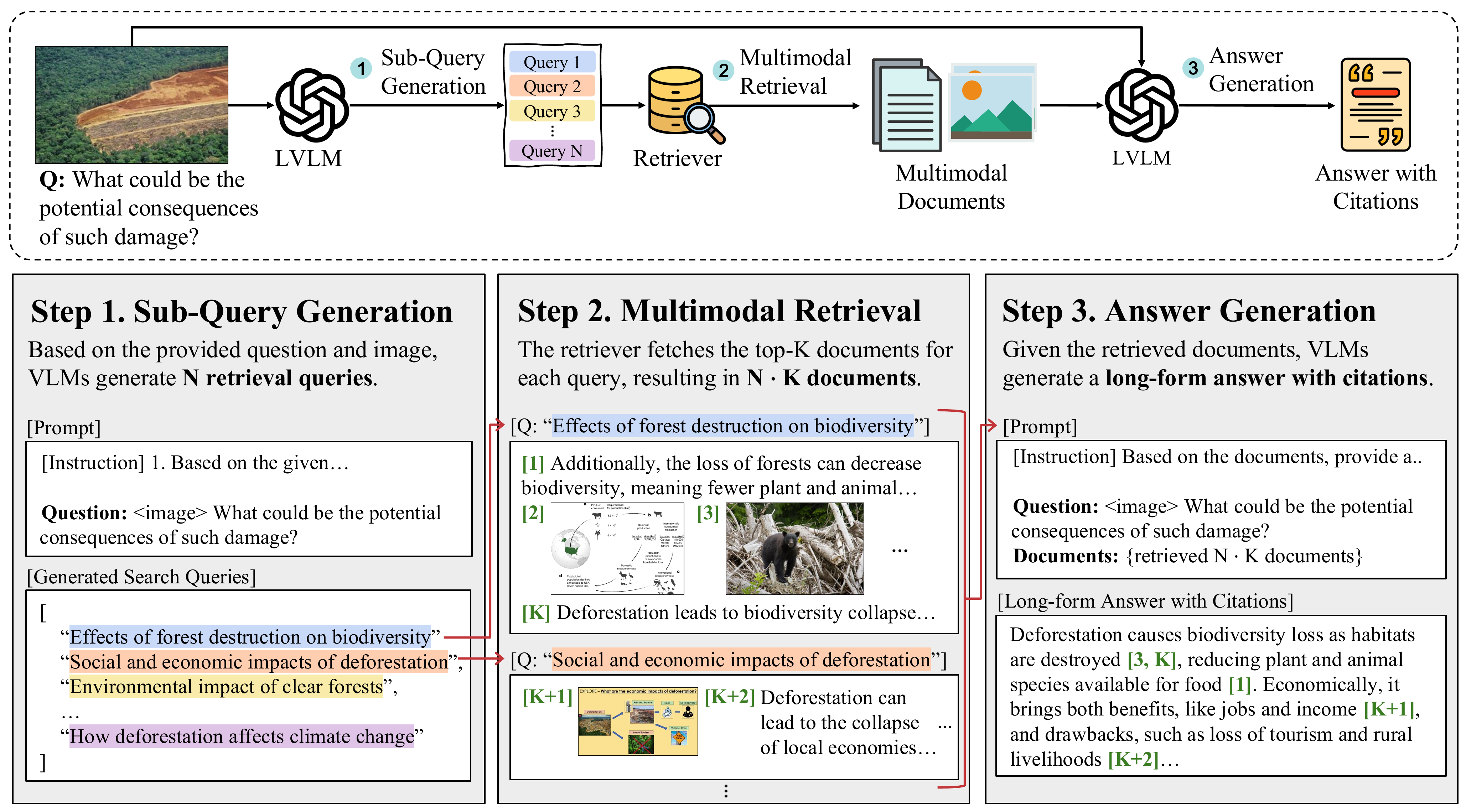}
\caption{An illustration of the task formulation in our benchmark.}
\label{fig:openended}
\end{figure*}

\section{Related Works}

\paragraph{Source attribution.}

Open-domain generative systems often produce plausible yet inaccurate content, known as hallucinations~\cite{ye2023cognitive, zhang2023siren}. To mitigate this, attribution techniques have been introduced, enabling models to provide supporting evidence in the form of citations to improve verifiability~\cite{gao2023enabling, li2024improving, sun2023towards, slobodkin2024attribute, huang2024training, li2024think}. However, these methods primarily address text-based inputs, with limited attention to the multimodal settings. We bridge this gap by adapting attribution approaches to multimodal contexts and identifying key factors for enhancing performance.

\paragraph{Long-form question answering (LFQA).} 
LFQA datasets are widely used in AQA tasks, as long-form responses are prone to including inaccurate information~\cite{liu2023evaluating}. ELI5~\cite{fan2019eli5} and HowSumm~\cite{boni2021howsumm} are popular LFQA datasets based on Reddit and WikiHow, respectively, and are evaluated by reference-based metrics such as ROUGE~\cite{lin2004rouge} and BLEU~\cite{papineni2002bleu}. To better address the open-ended nature of LFQA, LongFact~\cite{wei2024long} verifies atomic claims via web search, while RAG-QA Arena~\cite{han2024rag} employs pairwise preference evaluations. However, most existing datasets are still limited to text-only inputs, overlooking the role of multimodality. 
VizWiz-LF~\cite{huh2024long} introduces a long-form VQA task aimed at describing image content to blind or low-vision users. However, it does not consider grounding answers in external knowledge. 
M2RAG~\cite{ma2024multi} and MRAMB-Bench~\cite{yu2025mramg} introduce multimodal retrieval-augmented generation to produce multimodal answers. However, the images included in these answers serve as part of the response rather than as evidence supporting the statements.
Our work focuses on open-domain long-form VQA, which requires grounding answers in retrieved multimodal documents.

\section{The MAVIS Benchmark}

\subsection{Task formulation}
\label{sec:task_format}
In our task, given a visual question consisting of a user question $\mathcal{Q}$ and an input image $\mathcal{I}$, a vision-language model (VLM) $\mathcal{M}$ generate a long-form answer supported by verifiable evidence. Specifically, as illustrated in Figure~\ref{fig:openended}, the task involves the following three steps:

\begin{enumerate}
    \item \textbf{Search query generation:} Given $\mathcal{Q}$ and $\mathcal{I}$, $\mathcal{M}$ generates a list of $N$ search queries $\mathcal{S} = \{s_1, \dots, s_N\}$.
    
    \item \textbf{Multimodal document retrieval:} The retriever $\mathcal{R}$ fetches the top-$K$ documents $\mathcal{D}_i = \{d_{i, j}\}_{j=1}^{K}$ for each search query $s_i$. In total, $N \cdot K$ documents are retrieved, denoted as $\mathcal{D} = \{D_1, \dots, D_N\}$. 
    
    \item \textbf{Answer generation with citations:} Given the retrieved $\mathcal{D}$, $\mathcal{M}$ generates a final long-form answer with citations, each of which is enclosed in square brackets (e.g., [1]).
\end{enumerate}



\subsection{Dataset collection process} 
As discussed in \S\ref{sec:task_format}, our benchmark involves two essential components: a visual question and multimodal supporting documents. We automatically (1) collected long-form VQA data from user forums, and (2) retrieved supporting documents for each answer in fact-level. Subsequently, we (3) manually annotated a test set for reliable evaluation. More details about the dataset collection can be found in Appendix~\ref{appendix:data_collection}.

\subsubsection{Step 1. Collection of long-form VQA data}
\label{sec:data_filtering}
\paragraph{Raw data.} We collect 6M posts containing both images and comments from Reddit pushshift dumps from 2005-06 to 2023-12. We filter them by three rules: the title must be a question starting with a question word, ending with a question mark, and comments must be long (over 500 characters, 3$+$ sentences). By treating titles as questions and comments as answers, we obtain 432,817 VQA instances.

\paragraph{Visual dependency.} We filter out instances that can be answered without an input image, following \citet{chen2024we}. We instruct LLMs to answer the question without the input image and evaluate their answers against the forum answers. We utilize four LLMs—GPT‑4o~\cite{hurst2024gpt}, LLaMA‑3.3‑70B~\cite{dubey2024llama}, Mixtral‑8x7B~\cite{jiang2024mixtral}, and Phi‑4‑14B~\cite{abdin2024phi}—and employ InternVL‑2.5‑38B~\cite{chen2024internvl} as the evaluator. VQA instances correctly answered by at least one LLM are removed, resulting in 157,586 VQA instances. The prompts used for both LLMs and evaluator models are provided in Appendix~\ref{appendix:inspector_evaluator}.

\subsubsection{Step 2. Collection of supporting multimodal documents}
\label{sec:supporting}
\paragraph{Atomic fact extraction.}
A long-form answer typically contains multiple pieces of information~\cite{min2023factscore, jing2023faithscore}. We extract atomic facts from the ground-truth answer using GPT-4.1 with more details in Appendix~\ref{appendix:factcheck}.  On average, each answer contains 4.3 atomic facts.

\paragraph{Collection of multimodal documents.}
To collect external multimodal documents grounding atomic facts, we considered two search sources: Google Programmable Search, which searches the entire web for relevant images, and the Colossal Clean Crawled Corpus (C4)~\cite{raffel2020exploring}, a collection of hundreds of gigabytes of English text scraped from the web. Using each atomic fact as a query, we retrieved the top five passages from each source.

\paragraph{Automatic document filtering.}
Each VQA instance initially contains an average of 43 multimodal documents (4.3 facts × 2 modalities × 5 documents), which may overwhelm annotators during the filtering process.
Hence, we automatically remove irrelevant documents using entailment models (EMs).
Table~\ref{tab:entailment} shows that Qwen3-8B~\cite{yang2025qwen3} and SkyworkVLReward-8B~\cite{2025skyworkvlrm} perform the best for text and image entailment, respectively.
Using these models, we retain 930K text documents and 500K image documents.
Further details are provided in Appendix~\ref{appendix:entailment}.

\subsubsection{Step 3. Human annotation}
\label{sec:human_annotation}
\input{Tables/groundedness}
\paragraph{Sampling.} We sample a subset from the automatically collected dataset for human annotation. First, to ensure multimodal knowledge is necessary for answering the visual question, we select instances containing more than two atomic facts supported by multimodal documents. Second, to mitigate domain bias in the test set, we perform balanced sampling across different domains, resulting in 1,000 VQA instances. Detailed information is provided in Appendix~\ref{appendix:domain}.

\paragraph{Criteria.} Annotators label the sampled instances based on two criteria: (1) Are the atomic facts relevant to the question? (2) Are the facts accurately supported by the documents? As a result, we find that 85.7\% of atomic facts are relevant to the question. Among the relevant atomic facts, 87.3\% are supported by their corresponding documents. Finally, we annotate 1K test instances, comprising 3K relevant atomic facts supported by 5.1K ground-truth multimodal documents. Notably, all of our experimental results are based on this human-annotated test set. Further details on the annotation including inter-annotator agreement are in Appendix~\ref{appendix:human_annotation}.


\subsection{Human verification process}
To demonstrate the effectiveness of our data construction process, we involve human verifiers to assess whether the questions in our dataset (1) genuinely seek information or advice, as opposed to merely sharing information, advertising, or making statements, and (2) whether an attached image is necessary to understand or answer the user’s question, or unnecessary or irrelevant. We compare 50 visual questions from our dataset with their corresponding raw Reddit posts, each binary-labeled by two MTurk workers, with average inter-annotator agreement (IAA, measured by Cohen's Kappa) of 0.74 and 0.68, respectively. 

As a result, 89\% of the questions in our dataset exhibit information-seeking intent, compared to 49\% of the raw posts. Regarding image dependency, 89\% of questions in our dataset depend on attached images, whereas 68\% of raw posts require one. This confirms that our data construction process effectively removes irrelevant instances from the raw sources.

\section{Evaluation Metrics}
\label{metric} 
We evaluate long-form answers based on three aspects: (1) informativeness, (2) groundedness, and (3) fluency. We use GPT-4.1 as the evaluator for groundedness and informativeness, and MAUVE~\cite{pillutla2021mauve} for fluency. Further details are provided in Appendix~\ref{appendix:evaluation}.

\subsection{Informativeness}
Long-form answers can be paragraph-length responses that should be helpful and comprehensive. However, they often contain a large amount of information, making binary judgments challenging~\cite{min2023factscore}. To address this, we define two sub-metrics for \textit{informativeness}. (1) \textbf{Completeness} measures how thoroughly the model’s answer $\mathcal{A}$ covers the necessary information from the GT answer $\mathcal{G}$, and (2) \textbf{Relevance} assess whether $\mathcal{A}$ contains only information relevant to the user question $\mathcal{Q}$ and the user-provided image $\mathcal{I}$.

\begin{enumerate}
    \item \textbf{Completeness.} For the essential information that the answer $\mathcal{A}$ should cover, we use atomic facts $\{g_1, \dots, g_m\}$, extracted from $\mathcal{G}$ and filtered by human annotators. Given a fact $g_j$ and the model's answer $\mathcal{A}$, the completeness score $c(g_j, \mathcal{A})$ measures how thoroughly $g_j$ is addressed by $\mathcal{A}$ as $\{1$: \textit{fully relevant}, $0.5$: \textit{partially relevant}, $0$: \textit{not relevant}$\}$. The final completeness score is the average of $c(g_j, \mathcal{A})$ across all $g_j$:
    \[
        \text{Completeness}(\mathcal{A}) = \frac{1}{m} \sum_{j=1}^{m} c(g_j, \mathcal{A}).
    \]

    \item \textbf{Relevance.} Models should not generate  excessive or irrelevant information to achieve high completeness. Given the model’s answer $\mathcal{A}=\{a_1, \ldots, a_n\}$, user question $\mathcal{Q}$, and input image $\mathcal{I}$, the relevance score $r(a_i, \mathcal{Q}, \mathcal{I})$ indicates how appropriately each answer sentence $a_i$ addresses $\mathcal{Q}$ and $\mathcal{I}$ as $\{1$: \textit{fully relevant}, $0.5$: \textit{partially relevant}, $0$: \textit{not relevant}$\}$. The final relevance score is computed as the average relevance across all $a_i$:
    \[
        \text{Relevance}(A) = \frac{1}{n} \sum_{i=1}^{n} r(a_i, \mathcal{Q}, \mathcal{I}).
    \]
\end{enumerate}

\subsection{Groundedness}
We evaluate citation quality in terms of answer groundedness, using two sub-metrics. (1) \textbf{Recall} measures whether the answer is fully supported by citations, and (2) \textbf{Precision} identifies redundant or irrelevant citations. To do this, we first pair each sentence $a_i$ in the model's answer $\mathcal{A}$ with its corresponding cited documents $\mathcal{C}_i$ based on citation numbers. Each $a_i$ is then assigned a supportedness score $s(\cdot, \cdot)$ based on how well $a_i$ is supported by the documents as $\{1$: \textit{fully relevant}, $0.5$: \textit{partially relevant}, $0$: \textit{not relevant}$\}$.

\begin{enumerate}
    \item \textbf{Recall.} We assess how well each answer sentence $a_i \in \mathcal{A}$ is supported by its cited documents $\mathcal{C}_i$. The recall score for $\mathcal{A}$ is the average of $s(a_i, \mathcal{C}_i)$ across all $a_i$:
    \[
        \text{Recall}(\mathcal{A}) = \frac{1}{n} \sum_{i=1}^{n} s(a_i, \mathcal{C}_i).
    \]

    \item \textbf{Precision.} We evaluate how relevant every cited document is with its answer. For each answer sentence $a_i$, let $\mathcal{C}_i = \{c_{i,1}, \dots, c_{i,m}\}$ denote the set of its cited documents. The precision score for the answer $\mathcal{A}$ is the average, across all sentences $a_i$, of the mean supportedness scores between $a_i$ and each citation in $\mathcal{C}_i$:
    \[
        \text{Precision}(\mathcal{A}) = \frac{1}{n} \sum_{i=1}^{n} \left( \frac{1}{|\mathcal{C}_i|} \sum_{c_{i, j} \in \mathcal{C}_i} s(a_i, c_{i, j}) \right).
    \]
\end{enumerate}

\subsection{Fluency}
To measure how fluent and human-like the model’s answer \( \mathcal{A} \) is,  we adopt MAUVE~\cite{pillutla2021mauve}, as done in \citep{gao2023enabling}.  
Fluency mainly serves as a sanity check, ensuring MAUVE scores remain sufficiently high.

\section{Experiments}
\label{sec:experiments}

\subsection{Models}
\label{sec:models}

\paragraph{Large vision language models.} We select four state-of-the-art LVLMs. For proprietary models, we use (1) Claude-3.5-Sonnet-20241022, and (2) GPT-4o-240806~\cite{hurst2024gpt}. For public models,
we use (3) LLaVa-OneVision-Qwen2-72b-ov-hf~\cite{li2024llava}, and (4) QwenVL-72B-Instruct~\cite{wang2024qwen2}. Implementation details are explained in Appendix~\ref{appendix:implementation}.

\paragraph{Multimodal retrievers.}
\input{Tables/automatic_retrieval_result} We build a large-scale multimodal database of 2.5M documents—1.4M from our data and 1.1M (389K images, 787K texts) from the WebQA corpus~\cite{chang2022webqa}. We evaluate several multimodal retrievers, including CLIP-DPR~\cite{liu2022universal}, UniVL-DR~\cite{liu2022universal}, MARVEL~\cite{zhou2023marvel}, CLIP-SF~\cite{wei2024uniir}, and MM-Embed~\cite{lin2024mm}. For each test instance, we generate four search queries using the aforementioned LVLMs and measure retrieval accuracy against the GT supporting documents. As shown in Table~\ref{tab:retrieval}, the modality-aware retriever MM-Embed shows the best performance using averaged query embeddings for both text and image queries. Thus, we adopt MM-Embed as our default retriever.

\input{Tables/single_retrieval_vanilla}
\input{Tables/single_retrieval_explore}

\subsection{Baselines}

\paragraph{Retrieval modalities.}
We evaluate three RAG settings based on the modality of the knowledge base: Text-RAG, Image-RAG, and Multi-RAG. In Text-RAG and Image-RAG, the retriever selects documents from the corresponding unimodal database. In Multi-RAG, the retriever selects documents from a combined text-image database.

\paragraph{Answer generation.} We explore three answer generation methods. Instructions are detailed in Appendix~\ref{appendix:task_instruction}.

\begin{itemize}
    \item \textbf{Vanilla prompting}: We prompt each model to generate answers with inline citations. This end-to-end approach enables the simultaneous generation of sentences and their corresponding citations.
    
    \item \textbf{Chain-of-Thought (CoT) prompting}: Previous studies~\cite{slobodkin2024attribute, berchansky2024cotar} adopt Chain-of-Thought (CoT) prompting~\cite{wei2022chain} to enhance the accuracy of attributions. We utilize a guided reasoning framework consisting of: (1) finding relevant documents from the given set, (2) extracting relevant information from each document, and (3) generating the final answer using the relevant information.

    \item \textbf{Knowledge Extraction (KE) step}: LVLMs may reference documents but fabricate interpretations to generate wrong plausible answers. To mitigate this, we introduce an additional step prior to answer generation, independently extracting knowledge from documents without using the user question $Q$ or input image $I$, thereby preventing potential bias introduced by these inputs. Given documents $\mathcal{D}$, we extract factual information $\tilde{\mathcal{D}} = \{ \mathcal{M}(\mathcal{I}_\text{extract}, d) \mid d \in \mathcal{D}\}$ using a knowledge extraction instruction $\mathcal{I}_\text{extract}$, and then the VLM $\mathcal{M}$ generates a final long-form answer with citations from $\tilde{\mathcal{D}}$, while groundedness is evaluated with the original documents.

\end{itemize}

\begin{figure*}[ht!]
\centering
\includegraphics[width=\linewidth]{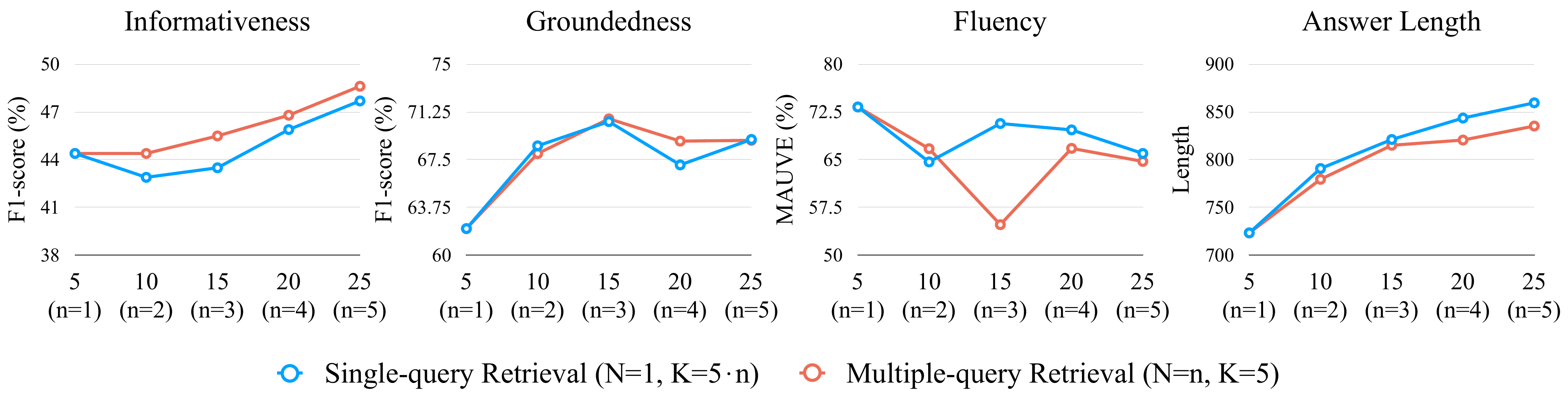}
\caption{Performance of Multi-RAG with GPT-4o using single-query ($N=1$) and multiple-query ($N=n$) retrieval. All baselines use Vanilla prompting. The $x$-axis indicates the total number of retrieved documents.} 
\label{fig:large_retrieval}
\end{figure*}

\subsection{Results for single-query retrieval}
We report automatic evaluation results in the single-query retrieval setting ($N = 1$), where the Top-5 documents ($K = 5$) are retrieved. Table~\ref{table:single_vanilla} compares performance across different retrieval modalities. Table~\ref{table:single_attribute} explores different answer generation methods in the Multi-RAG setting.

\paragraph{Utilizing multimodal documents improves informativeness and fluency.}
In Table~\ref{table:single_vanilla}, we compare retrieval modalities to identify the most effective knowledge modality for MAVIS. All models, except LLaVa-OneVision, demonstrate increased informativeness when using multimodal retrieval. For instance, Multi-RAG with GPT-4o achieves a 7.4\% higher F1-score in informativeness compared to Text-RAG, and a 4.2\% higher score than Image-RAG. Additionally, while all baselines show good fluency overall, Multi-RAG shows comparatively better fluency than uni-RAG, except for Qwen2.5VL.

\paragraph{LVLMs struggle to ground their answers in image documents.}
In Table~\ref{table:single_vanilla}, Image-RAG performs the lowest groundedness, with F1-scores ranging from 17.0\% to 53.1\%, whereas Text-RAG attains the highest groundedness with F1-scores from 66.7\% to 73.4\%. This indicates that LVLMs struggle to generate accurate citations when referencing image documents compared to text documents. Multi-RAG shows a 4.0\%–11.3\% decrease in groundedness relative to Text-RAG; this reduction is further analyzed in \S\ref{sec:multimodal_knowledge}.


\paragraph{Trade-off between informativeness and groundedness across prompting methods.}
In Table~\ref{table:single_attribute}, we compare two prompting methods—Vanilla and Chain-of-Thought (CoT)—in Multi-RAG, and observe a trade-off between informativeness and groundedness. For instance, CoT prompting with Qwen2.5VL increases informativeness by 6.3\% but decreases groundedness by 3.8\% compared to Vanilla prompting. Conversely, GPT-4o with CoT prompting increases groundedness by 7.2\% while decreasing informativeness by 11.7\% compared to Vanilla prompting. These results highlight the inherent difficulty of simultaneously improving informativeness and groundedness.

\paragraph{Knowledge extraction can mitigate this trade-off.}
As shown in Table~\ref{table:single_attribute}, knowledge extraction improves one metric (informativeness or groundedness) without sacrificing performance on the other. For GPT-4o with Vanilla prompting, after applying the knowledge extraction (KE) step, groundedness and informativeness increase by 4.3\% and 0.4\%, respectively. Similarly, GPT-4o with CoT prompting exhibits improvements of 4.2\% in informativeness and 0.7\% in groundedness after applying KE. Further analysis of the knowledge extraction step is provided in \S\ref{sec:multimodal_knowledge}.

\subsection{Results of retrieving more documents}

In Figure~\ref{fig:large_retrieval}, we present automatic evaluation results for retrieving more documents ($\geq 5$) using Vanilla prompting in Multi-RAG. We compare two baseline strategies while retrieving the same number of $(n \cdot k)$ documents. (1) Single-query retrieval: a single query ($N=1$) retrieves $K=n \cdot k$ documents, and (2) multiple-query retrieval: $N=n$ queries are generated, each retrieving $K=k$ documents.

\paragraph{Effects of retrieving more documents.}
Both informativeness and answer lengths generally increase with the number of retrieved documents. Specifically, retrieving 25 documents yields up to a 4.2\% increase in informativeness compared to retrieving 5 only. Groundedness also improves with more documents, peaking at 15 documents with a 8.4\% gain, beyond which additional documents offer no further benefit and slightly reduce performance. However, fluency shows a downward trend as more documents are retrieved, suggesting that integrating multiple documents may negatively impact the naturalness of model responses.

\paragraph{Effects of multiple-query retrieval.}
The multiple-query retrieval can incorporate diverse information. Thus, it achieves greater informativeness compared to single-query retrieval. However, multiple-query retrieval exhibits less stable fluency, indicating that increased document diversity may negatively affect response fluency.

\input{Tables/agreement}
\subsection{Results for human evaluation}
\label{human_eval}
We conduct a human study to validate automatic evaluation results with GPT-4.1. We collect 100 answers each from Qwen2.5-VL-72B and GPT-4o across three RAG frameworks (Image-RAG, Text-RAG, Multi-RAG) using Vanilla prompting, totaling 600 answers. Annotators rate answers on recall $s(a_i, \mathcal{C}_i)$, precision $s(a_i, \mathcal{C}_i)$, completeness $c(g_j, \mathcal{A})$, and relevance $r(a_i, \mathcal{Q}, \mathcal{I})$. As shown in Table~\ref{tab:corr_human}, Pearson correlation coefficients exceed 0.733 for all criteria, confirming GPT-4.1's reliability in the automatic evaluation of long-form answers. The annotation guidelines, inter-annotator agreement, and detailed results are provided in Appendix~\ref{appendix:human_eval}.

\begin{figure}[t!]
\centering
\includegraphics[width=\linewidth]{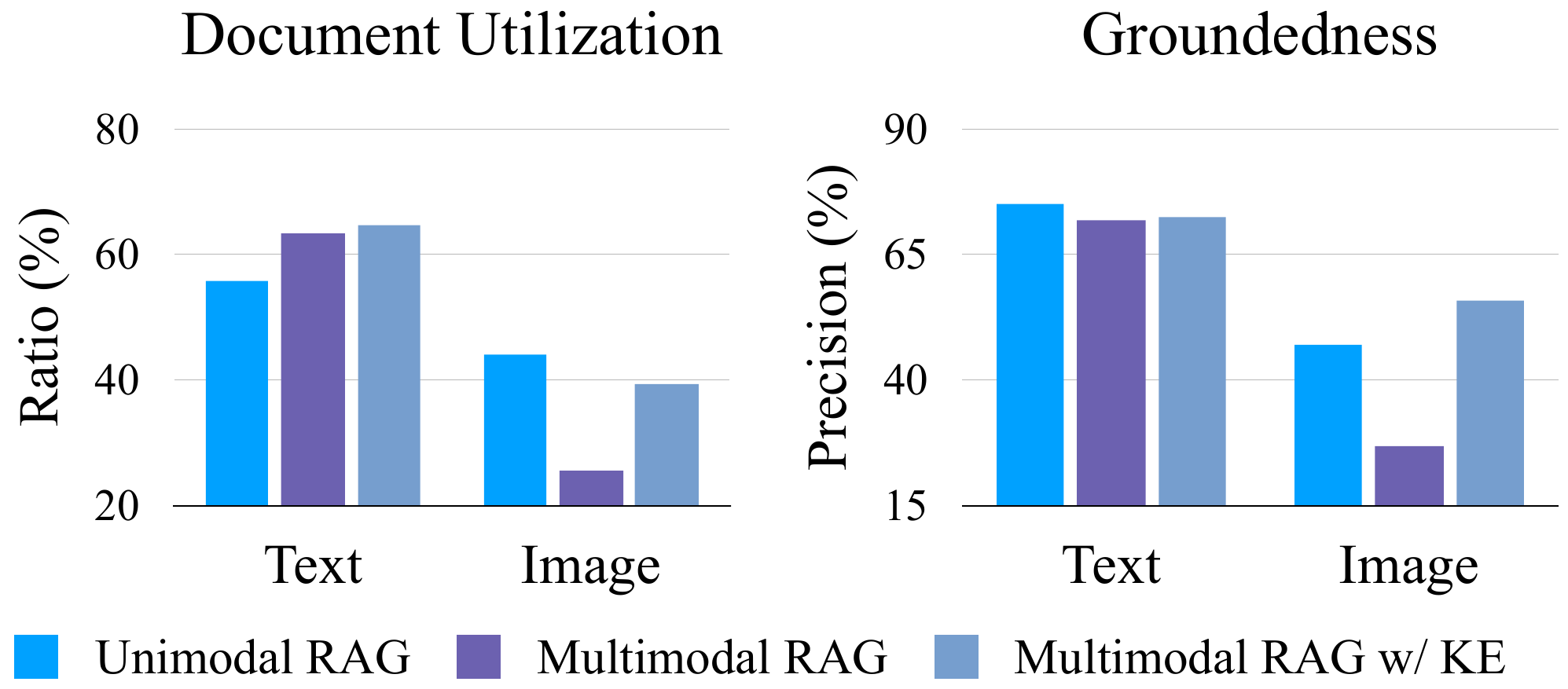}
\caption{Comparison of per-modality groundedness and document utilization between Unimodal-RAG (Text-RAG and Image-RAG), and Multimodal-RAG with and without knowledge extraction (KE). Each metric is averaged over the four LVLMs described in \S\ref{sec:models}.}
\label{fig:modality_precision}
\end{figure}

\subsection{Results for modality-specific evaluation}
\label{sec:multimodal_knowledge}

To investigate the capability of LVLMs to utilize documents for each modality, we conduct a modality-specific evaluation using two document-centric metrics: groundedness and the document utilization ratio. Groundedness is measured using precision rather than recall to assess the model's ability to utilize each document. The document utilization ratio per modality is defined as the average ratio of the number of used documents to that of retrieved documents.

\paragraph{How do LVLMs handle multimodal knowledge?} In Figure~\ref{fig:modality_precision}, we analyze the changes of each modality's contribution from a unimodal to a multimodal setting by comparing Multi-RAG with Uni-RAG (Text-RAG and Image-RAG). In Multi-RAG, text groundedness slightly decreases from 75.2\% to 71.9\%, whereas image groundedness substantially drops from 46.9\% to 26.8\%. Additionally, text utilization increases from 55.9\% to 63.5\%, while image utilization notably decreases from 44.0\% to 25.5\%. These results indicate that compared to Uni-RAG, Multi-RAG tends to primarily rely more on textual documents when generating answers, resulting in less attention to image documents compared to Uni-RAG.

\paragraph{Which modality would benefit from knowledge extraction?}
We propose a knowledge extraction step to mitigate the issue of LVLMs that fabricate to generate plausible answers. To identify which document modality benefits the most, we compare Multi-RAG and Multi-RAG w/ KE baselines. Significant improvements are observed for image documents, with groundedness rising from 26.8\% to 56.0\%, and document utilization from 25.5\% to 39.3\%. This indicates that LVLMs frequently fabricate information from image documents, and knowledge extraction effectively addresses this issue by mitigating contextual biases during interpretation.

\begin{figure}[t!]
\centering
\includegraphics[width=\linewidth]{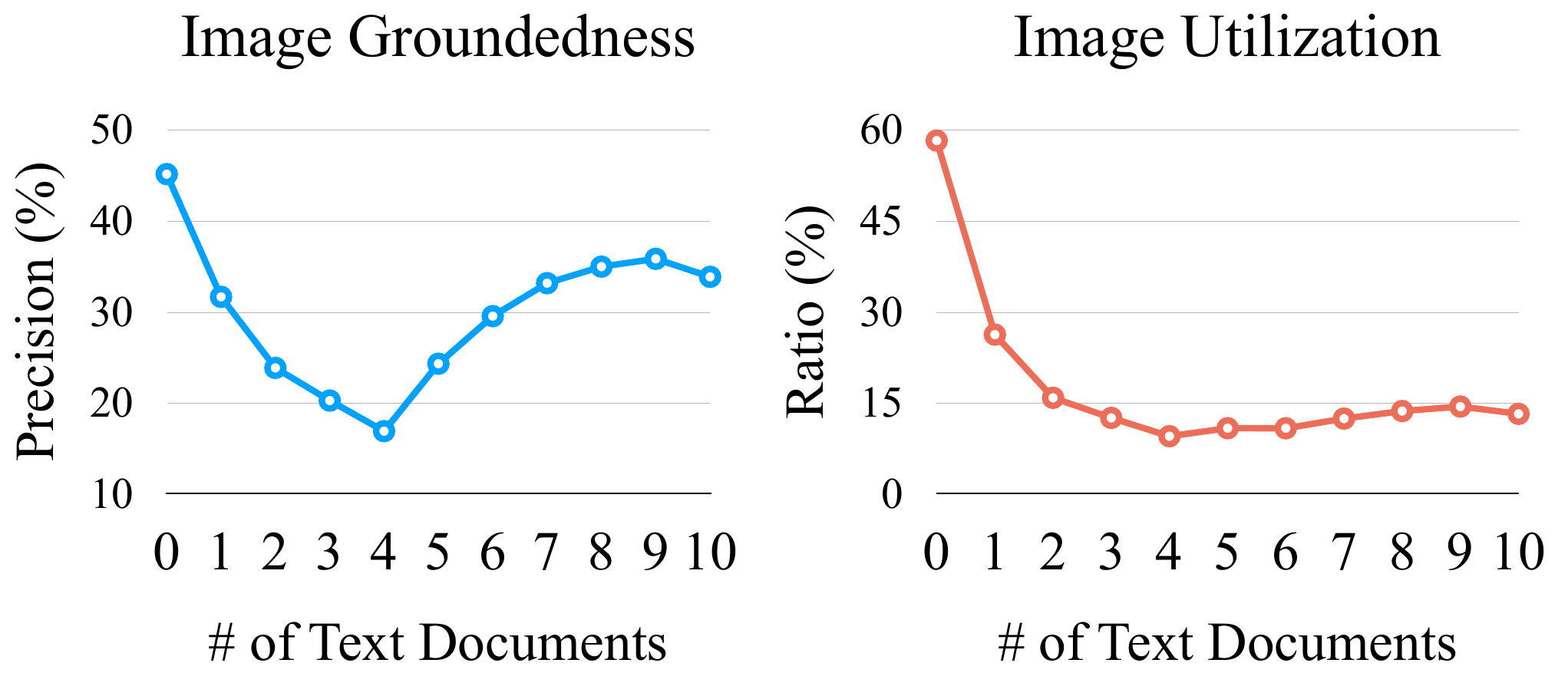}
\caption{Effect of adding text documents on image attention in GPT-4o. We fix the number of image documents at 5 in a single-query setting ($N=1$), and the x-axis represents the number of additional retrieved text documents. Documents are given in random order.}
\label{fig:controlled_experiment}
\end{figure}
\paragraph{To what extent do text documents affect attention to image documents?}
We found that LVLMs’ reliance on text reduces their attention to image documents. To investigate this further, we conducted a controlled experiment, shown in Figure~\ref{fig:controlled_experiment}, to examine how the number of text documents affects attention to images. First, when a single text document is added—compared to when no text documents are present—image utilization drops significantly, from 58.2\% to 26.25\%. This suggests that even one text document, which is small relative to the number of images, can cause GPT-4o to generate more text-focused responses. Second, when the number of text documents exceeds four, image utilization remains roughly constant at around 15\%, but image groundedness increases substantially, from 16.9\% to 33.85\%. These results indicate that, as recent studies~\cite{deng2025words, wu2025language} have reported, LVLMs exhibit text dominance and underutilize image documents; however, increasing the number of text documents beyond a certain point improves the accuracy of image document grounding.

\section{Future Directions}
In our experiments, we focus on prompting LVLMs without updating their model weights. By releasing sentence-level multimodal attribution data, we leave the exploration of optimizing source attribution in multimodal scenarios. While our work focuses on text and image documents, models could also leverage other modalities, such as video or audio, to ground their answers. For instance, video is particularly suitable for understanding dynamic events, motion, or sequences. Extending the dataset to include diverse modalities is an important direction for future work.

\section{Conclusion}

We introduce MAVIS, a benchmark for evaluating visual question answering with multimodal attribution. It includes 157K automatically generated instances and 1K manually annotated for evaluation. Experiments show that multimodal RAG yields more fluent, informative answers than unimodal RAG, but also reveals key limitations. LVLMs often over-rely on text, leading to weaker visual grounding. We also found a trade-off between informativeness and groundedness: higher grounding reduces informativeness, and vice versa. Our analysis highlights that stronger contextual bias with image documents than text, suggesting that an explicit knowledge extraction step can help mitigate this issue.

\section*{Acknowledgements}
We thank the anonymous reviewers for their valuable feedback.
This work was supported by 
Institute of Information \& Communications Technology Planning \& Evaluation (IITP) grant funded by the Korea government (MSIT) (No.~RS-2019-II191082, SW StarLab, No.~RS-2022-II220156, Fundamental research on continual meta-learning for quality enhancement of casual videos and their 3D metaverse transformation, No.~RS-2025-02263841, Development of a Real-time Multimodal Framework for Comprehensive Deepfake Detection Incorporating Common Sense Error Analysis, and No. RS-2021-II211343, Artificial Intelligence Graduate School Program, Seoul National University), 
and Sovereign AI Foundation Model Project (Data Track), organized by the Ministry of Science and ICT (MSIT) and supported by the National Information Society Agency (NIA) (2025-AI Data-wi43). 

\bibliography{aaai2026}

\clearpage
\input{supplementary-latex-2026}

\end{document}

%% file: Tables/data_compare.tex
\newcommand{\tmark}{\textcolor{yellow}{\ding{115}}}%
\newcolumntype{?}{!{\vrule width 1pt}}

\begin{table*}[t!]
    \centering
    \resizebox{\linewidth}{!}{
    \begin{NiceTabular}{lcccccccccc}
        \toprule
        & \multicolumn{3}{c}{\textbf{\# of Instance}} & \multicolumn{2}{c}{\textbf{Document}} & \multirow{2}{*}{\textbf{\makecell{Answer\\Length}}} & \multirow{2}{*}{\textbf{\makecell{Document\\Modality}}} & \multirow{2}{*}{\textbf{\makecell{Fact-level\\Citation}}} & \multirow{2}{*}{\textbf{\makecell{Task\\Formulation}}} \\
        \cmidrule(r){2-4} \cmidrule(r){5-6} 
        \textbf{Dataset} & \text{$Q$} & \text{$I$} & \text{$A$} & \text{$D_T/Q$} & \text{$D_I/Q$} &  &  &  &  \\
        \midrule
        ELI5 (\citeyear{fan2019eli5}) & 272,000 & 0 & 272,000 & 1.0 & 0 & 130.6 & Text & \xmark & Long-form QA \\\midrule
        AquaMuse (\citeyear{kulkarni2020aquamuse}) & 5,519 & 0 & 5,519 & 6.0 & 0 & 105.9 & Text & \xmark & Summarization \\\midrule
        HowSumm (\citeyear{boni2021howsumm}) & 95,469 & 0 & 95,469 & 10.1 & 0 & 150.2 & Text & \xmark & Summarization \\\midrule
        WikihowQA (\citeyear{bolotova2023wikihowqa}) & 11,746 & 0 & 11,746 & 6.3 & 0 & 149.3 & Text & \xmark & Long-form QA \\\midrule
        LFRQA (\citeyear{han2024rag}) & 26,907 & 0 & 26,907 & 3.0 & 0 & 76.3 & Text & \cmark & \makecell{Long-form QA \&\\Text Retrieval} \\\midrule
        LONGFACT (\citeyear{wei2024long}) & 2,280 & 0 & 0 & 0 & 0 & - & - & \xmark & Long-form QA \\\midrule
        VizWiz–LF (\citeyear{huh2024long}) & 600 & 600 & 4,200 & 0 & 0 & 41.2 & - & \xmark & Long-form VQA \\\midrule

        M2RAG  (\citeyear{ma2024multi}) & 750 & 0 & 0 & 0$^{\dagger}$ & 0$^{\dagger}$ & - & Text, Image & \xmark & \makecell{Multimodal Generation} \\\midrule
        
        MRAMB-Bench  (\citeyear{yu2025mramg}) & 4,800 &  7,340 & 4,800 & 1.2 & 1.8 & 134.8 & Text, Image & \xmark & \makecell{Multimodal Generation} \\\midrule
        
        \text{MAVIS (Ours)} & \makecell{81,173\\(991)} & \makecell{67,140\\(931)} & \makecell{157,586\\(1,000)} & \makecell{5.9\\(3.3)} & \makecell{3.1\\(2.0)} & 76.5 & Text, Image & \cmark & \makecell{Long-form VQA \&\\Multimodal Attribution} \\
        \bottomrule
    \end{NiceTabular}}
    \caption{Comparison of long-form question answering (LFQA) benchmarks. \( Q \), \( I \), and \( A \) denote the number of unique questions, images, and answers, respectively. \( D_T/Q \) and \( D_I/Q \) are the average number of text and image documents per question. The number in parentheses in our dataset indicates the statistics of the human-annotated set. The answer length is the average word count of answers. Fact-level citation indicates whether supporting documents are available for each verifiable fact. The dash (–) indicates that the corresponding feature is not covered in that benchmark. ${\dagger}$ denotes the absence of an annotated document for each question; however, an external knowledge base is utilized for retrieval.} 
    \vspace{-2mm}
    \label{tab:dataset_comparison}
\end{table*}

%% file: Tables/groundedness.tex
\begin{table}[t!]
    \centering
    \small
    \begin{NiceTabular}{llcc}
        \toprule
        \multirow{2}{*}{\textbf{Type}} & \multirow{2}{*}{\makecell{\textbf{Model}}} & \multicolumn{2}{c}{\textbf{F1-score}}\\\cmidrule(r){3-4}
        & & \textbf{Text} & \textbf{Image} \\
        \midrule
        \multirow{3}{*}{Textual}  & NLIDeBERTaV3-184M~\citeyearpar{he2021debertav3}  & 41.4 & - \\
                                  & FlanT5Verifier-11B~\citeyearpar{sanyal2024machines}         & 40.7 & - \\
                                  & Qwen3-8B~\citeyearpar{yang2025qwen3}     & \textbf{53.7} & - \\
        \midrule
        \multirow{2}{*}{Visual}   & OFA-VE-470M~\citeyearpar{wang2022ofa} & - & 23.1\\
                                  & SkyworkVLReward-8B~\citeyearpar{2025skyworkvlrm}         & - & \textbf{50.1} \\
        \midrule
        \multirow{1}{*}{Multi}     & Qwen2VL-7B~\citeyearpar{wang2024qwen2} & 45.0 & 49.6\\
        \bottomrule
    \end{NiceTabular}
    \caption{Model performances in verifying the groundedness of each sentence on the retrieved documents across different modalities. The F-1 scores are measured by a small set of ground truths that the authors manually label whether each document supports the corresponding fact. Due to computational constraints, we use about 11B open-source models.}
    \label{tab:entailment}
\end{table}

%% file: Tables/automatic_retrieval_result.tex
\begin{table}[!t]
\centering
\resizebox{\linewidth}{!}{
\begin{tabular}{lcc}
\toprule
\text{Retriever} & \text{NDCG@10} & \text{Recall@100} \\
\midrule\midrule
\multicolumn{3}{c}{Fine-tuned on WebQA~\cite{chang2022webqa}}\\\midrule
CLIP-DPR & 0.1567 & 0.4355\\
UniVL-DR & 0.1136 & 0.3244 \\
MARVEL-DPR & 0.1292 & 0.4188\\
MARVEL-ANCE & 0.1322&0.3948\\\midrule
\multicolumn{3}{c}{Fine-tuned on ClueWeb~\cite{overwijk2022clueweb22}}\\\midrule
MARVEL-DPR & 0.1098 & 0.4357 \\
MARVEL-ANCE & 0.1460 & 0.4398 \\\midrule
\multicolumn{3}{c}{Fine-tuned on M-BEIR~\cite{wei2024uniir}}\\\midrule
MM-Embed \\
+ text-seeking query & \text{0.2216} & \text{0.5909} \\
+ image-seeking query & \textul{0.2217} & \textul{0.6074} \\
+ averaged query embedding & \textbf{0.2565} & \textbf{0.6977} \\
\bottomrule
\end{tabular}}
\caption{Multimodal retrieval performance on the human annotated test set.}
\label{tab:retrieval}
\vspace{-3mm}
\end{table}

%% file: Tables/single_retrieval_vanilla.tex
\definecolor{graycustom}{HTML}{2f2f2f}

\begin{table*}[!t]
\centering
\small
\resizebox{\linewidth}{!}{
\begin{tabular}{lccc ccc c cc cc}
\toprule
& \multicolumn{7}{c}{Evaluation Metrics (\%)} & \multicolumn{4}{c}{Statistics} \\ \cmidrule(lr){2-8} \cmidrule(lr){9-12}
 & \multicolumn{3}{c}{\text{Informativeness}} & \multicolumn{3}{c}{\text{Groundedness}} & \text{Fluency} & \multicolumn{2}{c}{\text{Retrieved}} & \multicolumn{2}{c}{\text{Utilized}} \\\cmidrule(lr){2-4} \cmidrule(lr){5-7} \cmidrule(lr){8-8} \cmidrule(lr){9-10} \cmidrule(lr){11-12}
 & F1-score & Complete & Relevant & F1-score & Recall & Precision & MAUVE & Text & Image & Text & Image \\
\midrule

\textbf{\text{LLaVa-OneVision}} &  &  &  &  &  &  &  &  &  & & \\
\quad + Text-RAG & 31.9 & 26.0 & 78.0 & \textbf{66.7} & \textbf{70.3} & \textbf{65.5} & 80.4 & 5.0 & 0.0 & 3.2 & 0.0 \\
\quad + Image-RAG & \textbf{36.2} & \textbf{28.2} & \textbf{87.0} & 17.0 & 19.9 & 20.3 & 85.6  & 0.0 & 5.0 & 0.0 & 2.8 \\
\quad + Multi-RAG & 33.3 & 27.6 & 81.5 & 62.7 & 66.1 & 61.6 & \textbf{88.7} & 3.3 & 1.7 & 2.2 & 0.3 \\

\midrule

\textbf{\text{Qwen2.5VL}} &  &  &  &  &  &  &  \\
    \quad + Text-RAG & 30.7 & 23.4 & \textbf{77.8} & \textbf{77.9} & \textbf{79.8} & \textbf{77.0} & 60.1 & 5.0 & 0.0 & 3.4 & 0.0 \\
\quad + Image-RAG & 30.7 & 24.2 & 75.1 & 49.2 & 50.0 & 54.0 & \textbf{81.5} & 0.0 & 5.0 & 0.0 & 3.0 \\
\quad + Multi-RAG & \textbf{32.3} & \textbf{25.4} & 77.3 & 72.0 & 72.5 & 73.9 & 62.8 & 2.8 & 2.2 & 2.0 & 1.1 \\

\midrule

\textbf{\text{Claude-3.5-Sonnet}} &  &  &  &  &  &  &  \\
\quad + Text-RAG & 29.6 & 24.0 & 68.6 & \textbf{79.1} & \textbf{80.9} & \textbf{78.1} & 69.7 & 5.0 & 0.0 & 3.5 & 0.0 \\
\quad + Image-RAG & 32.3 & 28.5 & 68.9 & 53.1 & 52.2 & 57.7 & 71.5 & 0.0 & 5.0 & 0.0 & 3.2 \\
\quad + Multi-RAG & \textbf{35.4} & \textbf{29.3} & \textbf{75.6} & 74.7 & 75.2 & 76.0 & \textbf{72.9} & 2.5 & 2.5 & 2.1 & 1.1 \\

\midrule

\textbf{\text{GPT-4o}} &  &  &  &  &  &  &  \\
\quad + Text-RAG & 37.0 & 30.4 & 82.7 & \textbf{73.4} & \textbf{76.9} & \textbf{71.3} & 69.7 & 5.0 & 0.0 & 3.1 & 0.0 \\
\quad + Image-RAG & 40.2 & 38.0 & \textbf{87.4} & 37.1 & 38.4 & 45.4 & 71.5 & 0.0 & 5.0 & 0.0 & 2.7 \\
\quad + Multi-RAG & \textbf{44.4} & \textbf{42.5} & 85.4 & 62.1 & 65.0 & 63.2 & \textbf{73.3} & 2.9 & 2.1 & 2.1 & 0.6 \\

\bottomrule
\end{tabular}
}
\caption{Performance of LVLMs over modalities of knowledge base. Each baseline uses Vanilla prompting under the single retrieval setting ($N=1, K=5$). \textbf{Bold numbers} indicate the best performance. We calculate the F1-score for groundedness and informativeness as representative values.} 
\label{table:single_vanilla}
\vspace{-3mm}
\end{table*}

%% file: Tables/single_retrieval_explore.tex
\definecolor{graycustom}{HTML}{2f2f2f}

\begin{table}[!t]
\centering
\renewcommand{\arraystretch}{1.1}
\resizebox{\linewidth}{!}{
\begin{tabular}{lccc}
\toprule
 & \multicolumn{1}{c}{Inf.} & \multicolumn{1}{c}{Grd.} & Flu.\\\cmidrule{2-4}
Method & \multicolumn{1}{c}{F1-score} & \multicolumn{1}{c}{F1-score} & \multicolumn{1}{c}{MAUVE} \\
\midrule

\textbf{LLaVa-OneVision} & & & \\
\quad + Vanilla & 33.3 & 62.7 & \textbf{88.7} \\
\quad + Vanilla + KE (\textbf{Ours})  & \textbf{35.2} & \textbf{63.0} & 79.8 \\\midrule
\textbf{Qwen2.5VL} & & & \\
\quad + Vanilla & 32.3 & \textbf{72.0} & 50.8 \\
\quad + Vanilla + KE (\textbf{Ours})   &  34.9  &  \textbf{72.0} & 65.9 \\
\quad + CoT   & 38.6 & 68.2 & \textbf{72.4} \\
\quad + CoT + KE (\textbf{Ours})  & \textbf{39.0} & 70.4 & 69.1 \\\midrule

\textbf{Claude-3.5-Sonnet} & & & \\
\quad + Vanilla & 35.4 & 74.7 & 72.9 \\
\quad + Vanilla + KE (\textbf{Ours}) & 37.5 & \textbf{75.1}  & \textbf{96.2} \\
\quad + CoT   & 39.6 & 64.6 & 63.2 \\
\quad + CoT + KE (\textbf{Ours})  & \textbf{40.4} & 65.0  & 94.5 \\\midrule

\textbf{GPT-4o} & & & \\
\quad + Vanilla & 44.4 & 62.1 & 73.3 \\
\quad + Vanilla + KE (\textbf{Ours})  & \textbf{44.8} & 66.4 & 70.1 \\
\quad + CoT   & 32.7 & 69.3 & 79.9 \\
\quad + CoT + KE (\textbf{Ours})  & 36.5 & \textbf{70.0} & \textbf{81.2} \\

\bottomrule
\end{tabular}
}

\caption{Performance of LVLMs for each answer generation method under single-query retrieval setting ($N=1$, $K=5$). All baselines use the Multi-RAG setting. \textbf{Bold numbers} indicate the best performance.}

\label{table:single_attribute}
\vspace{-3mm}
\end{table}

%% file: Tables/agreement.tex
\begin{table}[t!]
\centering
\begin{tabular}{llc}
\toprule
Metric & Sub-metric & Pearson Corr.\\
\midrule
\multirow{2}{*}{Groundedness} & Recall & 0.903 \\
& Precision & 0.803\\\midrule
\multirow{2}{*}{Informativeness} & Completeness & 0.733 \\
& Relevance & 0.855\\
\bottomrule
\end{tabular}
\caption{Pearson correlation between human and GPT-4.1 scores for each metric.}
\vspace{-2mm}
\label{tab:corr_human}
\end{table}

%% file: supplementary-latex-2026.tex
\appendix
\label{sec:appendix}
\section{Data Collection Details}
\label{appendix:data_collection}
\subsection{Instruction for Inspector and Evaluator}
\label{appendix:inspector_evaluator}
\input{Tables/prompt_inspector}
\input{Tables/prompt_evaluator}
The instructions for the inspectors and the evaluator are shown in Table~\ref{tab:prompt_inspector} and Table~\ref{tab:prompt_evaluator}.

\subsection{Atomic Facts Extraction}
\label{appendix:factcheck}
\input{Tables/prompt_claim}
\input{Tables/prompt_claim_filter}
We use an LVLM-based approach to extract atomic facts with GPT-4.1. First, we split the answer into individual sentences using the NLTK sentence tokenizer~\cite{loper2002nltk}. Each sentence is then processed using the prompt in Table~\ref{tab:prompt_claim} to extract atomic facts. After extraction, we filter the facts using the prompt in Table~\ref{tab:prompt_claim_filter}, based on two criteria: (1) whether the fact is relevant to the question, and (2) whether it was provided by the questioner. Irrelevant facts are discarded. If a fact was explicitly stated by the questioner, it is treated as non-verifiable and removed from consideration.

\subsection{Entailment Model}
\label{appendix:entailment}

To identify the groundedness of each statement with respect to the corresponding document, we treat the document as the premise and the statement as the hypothesis. If an entailment model outputs the label ``entailment,'' we consider the statement to be grounded. We use different entailment models depending on the modality of the document.

\paragraph{Textual Entailment Models.}
We consider the following textual entailment models: \text{NLIDeBERTaV3-184M}~\cite{he2021debertav3}, \text{FlanT5Verifier-11B}~\cite{sanyal2024machines}, and \text{Qwen3-8B}~\cite{yang2025qwen3}.

For \text{NLIDeBERTaV3-184M}, we use the \texttt{text-classification} pipeline from the \texttt{transformers} library~\cite{wolf2019huggingface}. The model classifies input into one of three labels: \texttt{entailment}, \texttt{neutral}, or \texttt{contradiction}.

For \text{FlanT5Verifier-11B}, we use the following prompt template:
\begin{quote}
\texttt{Premise: \{premise\} Hypothesis: \{hypothesis\} Given the premise, is the hypothesis correct? Answer:}
\end{quote}
We then compute token probabilities for ``\texttt{Yes}'' and ``\texttt{No}''. If ``\texttt{Yes}'' has a higher probability, we classify the pair as \texttt{entailment}; otherwise, we classify it as \texttt{not entailment}.

For \text{Qwen3-8B}, we use a similar prompt:
\begin{quote}
\texttt{Premise: \{premise\} Hypothesis: \{hypothesis\} Given the premise, is the hypothesis correct? Respond in yes or no. Answer:}
\end{quote}
If the model outputs ``\texttt{yes}'', we treat the pair as \texttt{entailment}; otherwise, as \texttt{not entailment}.

\paragraph{Visual Entailment Models.}
We consider the following visual entailment models: \text{OFA-VE-470M}~\cite{wang2022ofa} and \text{SkyworkVLReward-8B}~\cite{2025skyworkvlrm}.

For \text{OFA-VE-470M}, we use the visual entailment pipeline, which is prompted with:
\begin{quote}
\texttt{Statement: \{statement\} Is this statement right according to the image? Please answer yes or no.}
\end{quote}
We classify the image-statement pair as \texttt{entailment} if the model outputs ``yes'', and \texttt{not entailment} otherwise.

For \text{SkyworkVLReward-8B}, we adopt a reward-based scoring approach. Given a premise image and a textual hypothesis, we prompt the model with:
\begin{quote}
\texttt{Determine whether the image entails the statement "\{statement\}". A. Yes. B. No.}
\end{quote}
We compute separate reward scores for the completions A. Yes. and B. No.'' The option with the higher reward score determines the final prediction.

\paragraph{Multimodal Entailment Model.}
We use \text{Qwen2VL-7B}~\cite{wang2024qwen2} as a multimodal entailment model. It is prompted as follows:
\begin{quote}
\texttt{Premise: \{premise\} Hypothesis: \{hypothesis\} Given the premise, is the hypothesis correct? Respond in yes or no. Answer:}
\end{quote}
If the model outputs ``yes'', we classify the image-hypothesis pair as \texttt{entailment}; otherwise, as \texttt{not entailment}.

\subsection{Domains of Subreddits}
\label{appendix:domain}

\input{Tables/domain}
Following \citet{yue2024mmmu}, we define six domains to ensure balanced question distribution across domains. The complete subreddit-to-domain mapping is provided in Table~\ref{tab:domains}.

\subsection{Human Annotation}
\label{appendix:human_annotation}

\begin{figure*}[t!]
\centering
\includegraphics[width=0.6\linewidth]{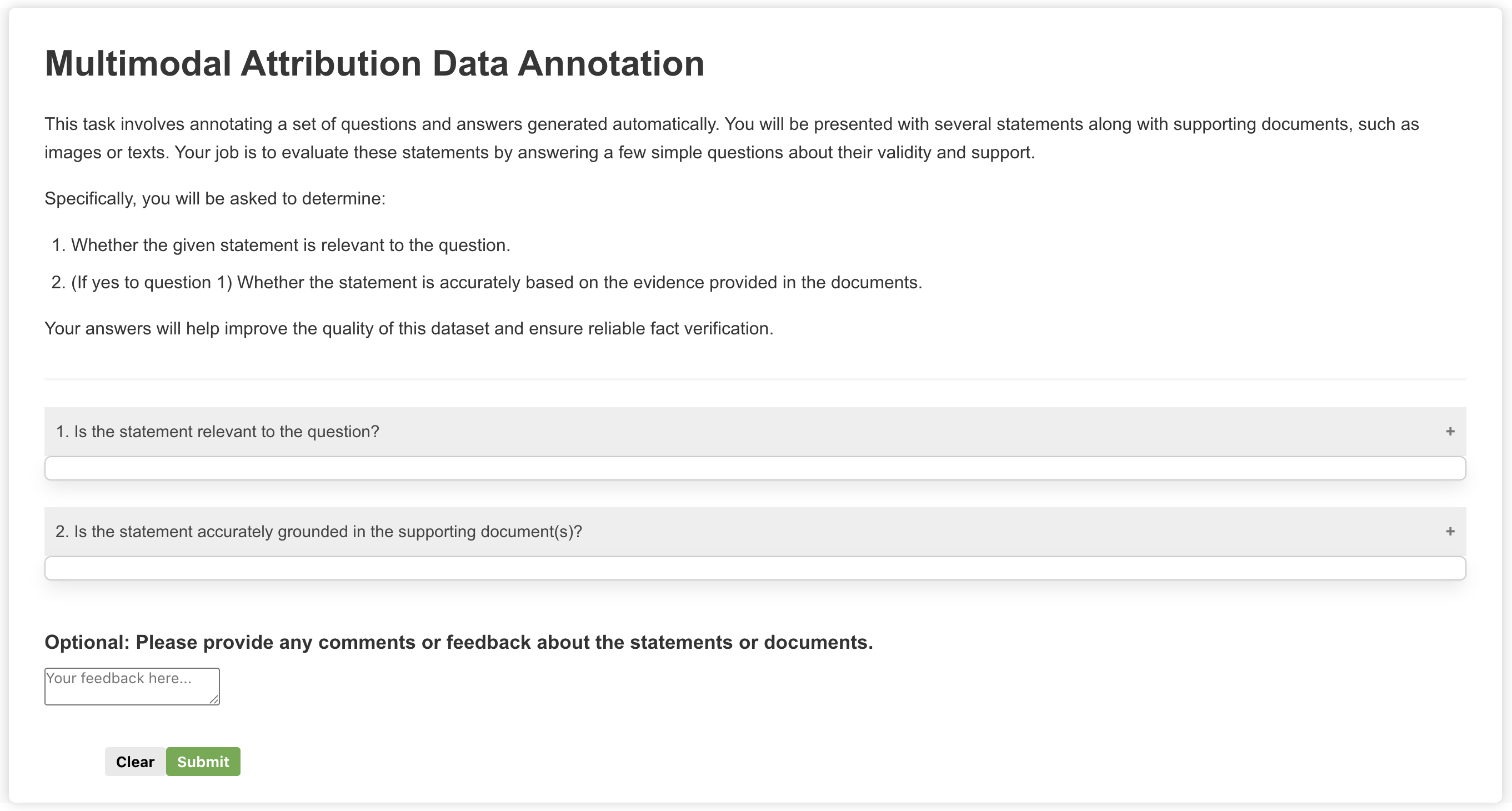}
\caption{Instructions provided for human evaluators to obtain.}
\label{fig:amt1}
\end{figure*}

\begin{figure*}[t!]
\centering
\includegraphics[width=0.6\linewidth]{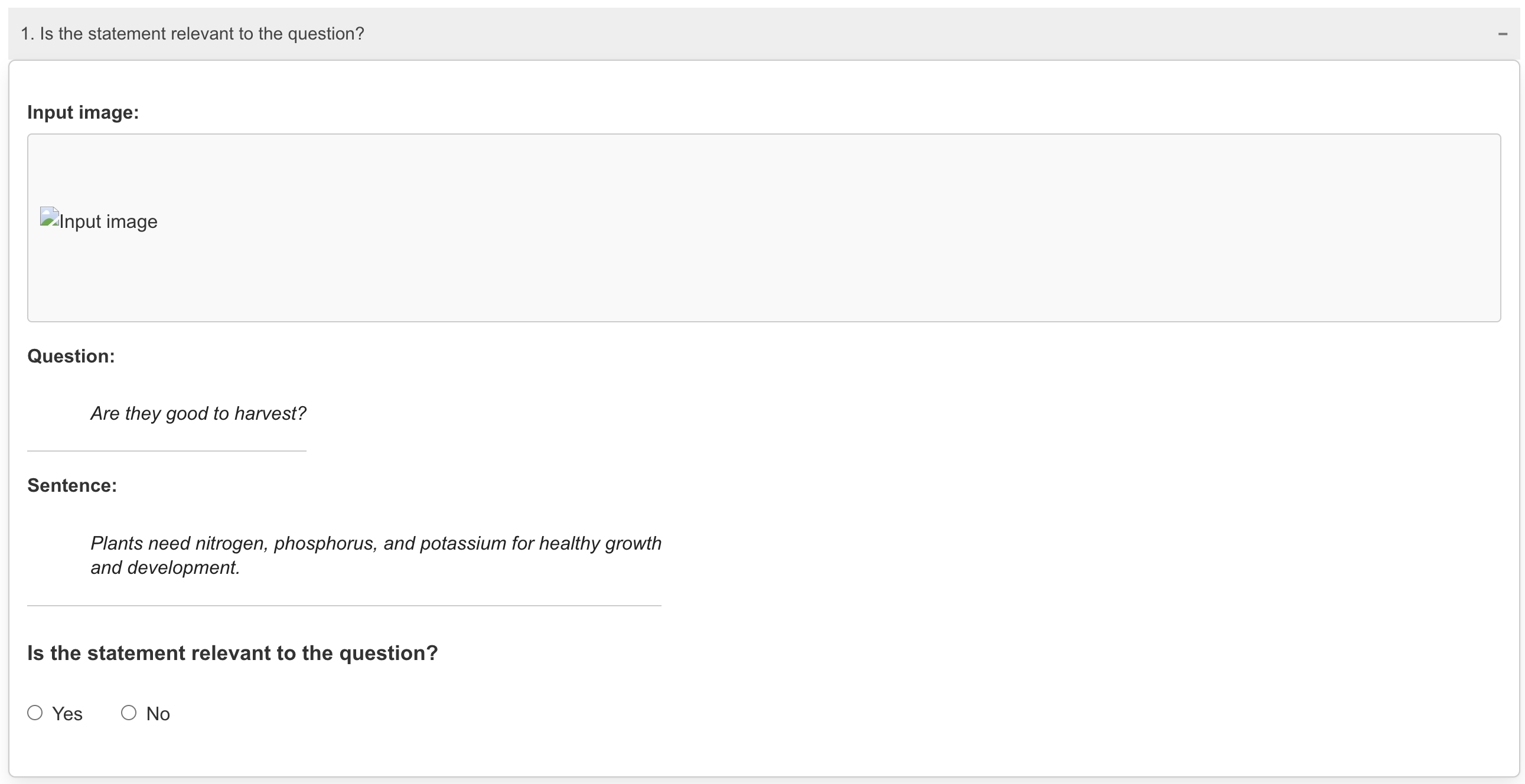}
\caption{Instructions template provided for human evaluators to obtain labels for fact relevance.}
\label{fig:amt2}
\end{figure*}

\begin{figure*}[t!]
\centering
\includegraphics[width=0.6\linewidth]{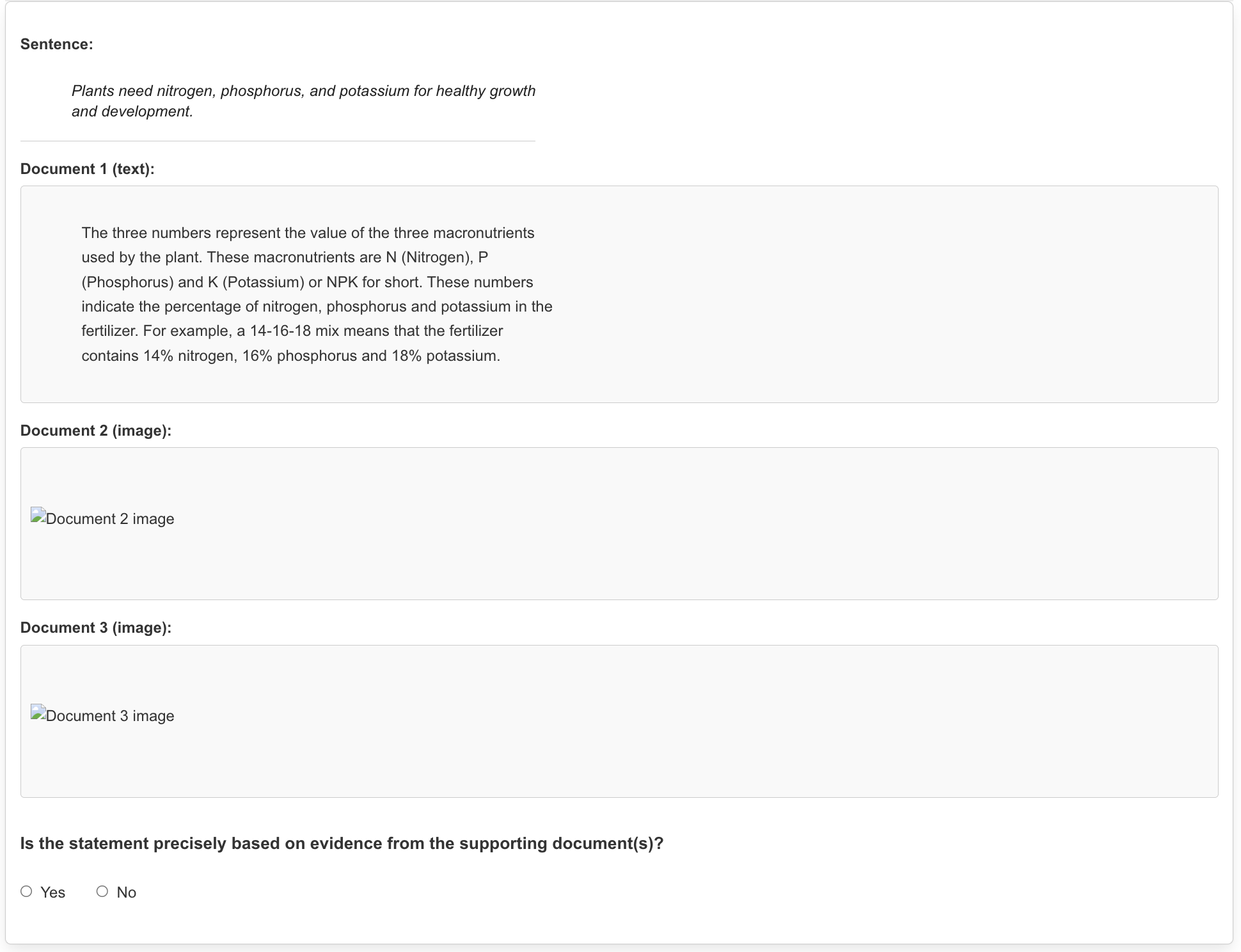}
\caption{Instructions provided for human evaluators to obtain labels for fact supportedness.}
\label{fig:amt3}
\end{figure*}

We hired data annotators via Amazon Mechanical Turk (MTurk). Five annotators were selected based on their performance in a qualification task designed to assess their ability to determine whether statements are accurately supported by the given documents. We required annotators to be from English-speaking countries (AU, CA, NZ, US, GB), have completed more than 10,000 HITs, and maintain a HIT approval rate above 98\%. The qualification task consisted of 10 examples (20 questions in total) and paid \$5.00 per qualification task. Each qualification task included three questions as illustrated in Figures~\ref{fig:amt1}, \ref{fig:amt2}, and \ref{fig:amt3}.

\begin{enumerate}
\item \textbf{Is the fact relevant to the question?}  
Determine whether the fact is relevant to the question and needs to be verified.

\item \textbf{Are the statements accurately grounded in the supporting documents?}
Verify if each statement is precisely grounded by referencing one or more external documents.
\end{enumerate}

We randomly extracted 1,000 QA instance pairs, consisting of almost 4,000 sentences (instances without supporting documents were removed).
Annotators labeled each pair using the two-question format described above. If annotators labeled question (1) as \textit{false} (sentence is irrelevant), they skipped question (2). For question (2), up to three documents were provided for each sentence. We measured inter-annotator agreement on a subset of 100 pairs in advance. Fleiss' $\kappa$ scores for binary classification were 0.75 for question (1) and 0.80 for question (2).

\section{Evaluation Details}
\label{appendix:evaluation}
\subsection{Model-based Evaluation}
\input{Tables/prompt_eval_groundedness}
\input{Tables/prompt_eval_completeness}
\input{Tables/prompt_eval_relevance}
In this section, we explain our instruction templates for automatic evaluation using GPT-4.1. For supportedness, see Table~\ref{tab:groundedness}. For completeness, see Table~\ref{tab:completeness}. For relevance, see Table~\ref{tab:relevance}.

\subsection{Human Evaluation}
\label{appendix:human_eval}


To verify the quality of the GPT-4.1-based automatic evaluation, we conducted a human evaluation with three graduate students, selected through a qualification task. This task involved rating 10 model-generated answers based on groundedness, completeness, and relevance. On average, participants spent approximately 100 minutes and were compensated \$15.00 for completing the qualification. The instructions for the human evaluation were identical to those used in the model-based evaluation, as shown in Tables~\ref{tab:groundedness}, \ref{tab:completeness}, and \ref{tab:relevance}.

After the qualification phase, each human evaluator assessed 200 model-generated answers with respect to supportedness, completeness, and relevance. Each answer was independently evaluated by two annotators, and we used the average of their scores as the final human evaluation result. Annotators were compensated \$1.50 per answer. Inter-annotator agreement (IAA)—excluding ratings by the authors—was measured using average Cohen’s $\kappa$, yielding 0.588 for supportedness, 0.648 for completeness, and 0.659 for relevance. Finally, we compared the averaged human evaluation scores with the model-based evaluation results.

\section{Experimental Details}

\subsection{Implementation Details}
\label{appendix:implementation}
For all models, we collect responses using nucleus sampling with temperature $\mathcal{T}=0.7$ and top-$p=0.95$, selecting the most likely sequence. The maximum number of generated tokens is set to 2048. Input images are rescaled such that the longer side—either width or height—does not exceed 512 pixels. For OpenAI and Anthropic models, we use an API-based approach to gather model responses. For LLaVA-OneVision and Qwen2.5-VL, we run inference locally using 8 $\times$ NVIDIA RTX A6000 GPUs with Intel(R) Xeon(R) Gold 6130 CPU @ 2.10GHz. FlashAttention-2 is used to accelerate attention computation.

\subsection{Details in Prompting}
\label{appendix:task_instruction}
\input{Tables/prompt_query_generation}
\input{Tables/prompt_vanilla}
\input{Tables/prompt_cot}
\input{Tables/prompt_agg}

In this section, we describe our task instruction templates.
For the search query generation prompt, refer to Table~\ref{tab:prompt_query}.
For the \textsc{Vanilla} prompt, see Table~\ref{tab:prompt_answer}.
For the \textsc{Chain-of-Thought} prompt, refer to Table~\ref{tab:prompt_cot}.
For the \textsc{Knowledge Extraction} step, refer to Table~\ref{tab:prompt_agg}.

%% file: Tables/prompt_inspector.tex
\begin{table*}[htbp]
\scriptsize
\centering
\begin{tabular}{@{}p{\linewidth}@{}}
\toprule
\textbf{Answer Generation Instruction for LLMs}\\
\midrule
Instruction:\\
1. Given a question, your task is to generate an answer. \\
2. Even if describing the image seems impossible without viewing it, you should predict the situation and describe it accordingly. \\
3. Only generate answer.\\
\\
Question: \{question\}\\
\bottomrule
\end{tabular}
    \caption{Instruction for LLMs without image.}
    \label{tab:prompt_inspector}
\end{table*}

%% file: Tables/prompt_evaluator.tex
\begin{table*}[htbp]
\scriptsize
\centering
\begin{tabular}{@{}p{\linewidth}@{}}
\toprule
\textbf{Answer Evaluation Instruction}\\
\midrule
Instructions:\\
1. Given an image, a question, a ground-truth answer, and a model response, your task is to evaluate whether the model response is ``right'' or ``wrong''. \\
2. Even if the model response differs from the gold answer, if the model appears to have correctly understood the image, label the response as ``right''. \\
\\
Question: $<$image$>$\{question\}\\
GT answer: \{gold\_answer\}\\
Model response: \{model\_response\}\\
\bottomrule
\end{tabular}
    \caption{Instruction for evaluator.}
    \label{tab:prompt_evaluator}
\end{table*}

%% file: Tables/prompt_claim.tex
\begin{table*}[htbp]
\scriptsize
\centering
\begin{tabular}{@{}p{\linewidth}@{}}
\toprule
\textbf{Atomic Facts Extraction Instruction}\\
\midrule
You and your partners are on a mission to fact-check a claim that may contain multiple subclaims that need to be verified. A sentence that needs to be verified is any statement or assertion that requires evidence or proof to support its accuracy or truthfulness. For example, “Titanic was first released in 1997” necessitates verification of the accuracy of its release date, whereas a claim like "Water is wet" does not warrant verification. Each subclaim is a simple, complete sentence with single point to be verified. Imagine yourself as an expert in processing sentences and extracting subclaims. Your task is to extract clear, unambiguous subclaims to check from the sentence, avoiding vague references like ’he,’ ’she,’ ’it,’ or ’this,’ and using complete names.\\
\\
To illustrate the task, here are some examples:\\
\{in-context examples\}\\
\\
Now, let’s return to your task. You are given the following sentence, please extract all subclaims that need to be checked.\\
\\
Sentence: \{sentence\}\\
Subclaims: \{extracted claims\}.\\
\bottomrule
\end{tabular}
    \caption{Instruction for atomic facts extraction following \citet{li2023self}.}
    \label{tab:prompt_claim}
\end{table*}

%% file: Tables/prompt_claim_filter.tex
\begin{table*}[htbp]
\scriptsize
\centering
\begin{tabular}{@{}p{\linewidth}@{}}
\toprule
\textbf{Atomic Facts Filtering Instructions}\\
\midrule
You and your partners are on a mission to determine whether a given fact is (1) relevant to the question and (2) provided by the questioner.\\
\quad - First, identify whether the fact is relevant to the question. Classify each fact as either ``Relevant'' or ``Irrelevant''.\\
\quad - Second, if the fact is classified as ``Relevant,'' determine whether it was provided by the questioner. If the fact is directly stated or implied by the questioner, label it as ``Provided''; otherwise, label it as ``Verifiable''.\\
\\
To illustrate the task, here are some examples:\\
\{in-context examples\}\\
\\
Now, let’s return to your task. You are given a question. Please classify the following fact according to the instructions above.\\
\\
Question: $<$image$>$\{question\}\\
Fact: \{fact\}.\\
\bottomrule
\end{tabular}
\caption{Instructions for filtering atomic facts.}
\label{tab:prompt_claim_filter}
\end{table*}

%% file: Tables/domain.tex
\begin{table*}[htbp]
\centering
\renewcommand{\arraystretch}{1.2}

\begin{tabular}{m{5cm}|p{11cm}}
\hline
\textbf{Domain} & \textbf{Subreddits} \\
\hline
Art \& Design & r/Art, r/Design, r/Filmmakers, r/GraphicDesign, r/Illustration, r/Music, r/architecture, r/femalefashionadvice, r/frugalmalefashion, r/malefashionadvice, r/musictheory, r/photocritique, r/vinyl, r/houseplants, r/gardening, r/HomeImprovement, r/woodworking \\
\hline
Business & r/personalfinance, r/Daytrading, r/StockMarket, r/wallstreetbets, r/Economics, r/algotrading \\
\hline
Science & r/science, r/nasa, r/math, r/chemistry, r/Physics, r/biology, r/neuroscience, r/Astronomy, r/AskPhysics, r/AskChemistry, r/askscience, r/AskStatistics, r/genetics, r/space, r/Futurology, r/askmath, r/learnmath, r/matheducation, r/datascience, r/astrophotography, r/environment, r/natureismetal, r/UFOs, r/singularity, r/theydidthemath, r/astrology \\
\hline
Health \& Medicine & r/Health, r/medicine, r/medical\_advice, r/AskMedical \\
\hline
Humanities \& Social Science & r/sociology, r/AskHistorians, r/AskHistory, r/history, r/classics, r/politics, r/PoliticalScience, r/linguistics, r/religion \\
\hline
Tech \& Engineering & r/SoftwareEngineering, r/engineering, r/ElectricalEngineering, r/AskEngineers, r/webdev, r/hardware, r/technology, r/pcmasterrace, r/gadgets, r/javascript, r/AskElectricians, r/AskElectronics, r/ChemicalEngineering, r/industrialengineering, r/buildapc, r/AskComputerScience, r/compsci, r/programming, r/techsupport, r/3Dprinting, r/raspberry\_pi, r/MechanicalEngineering, r/civilengineering, r/AerospaceEngineering, r/DIY, r/homeautomation, r/mac, r/hacking, r/aviation, r/dataisbeautiful \\
\hline
\end{tabular}

\caption{Entire subreddit-domain mapping.}
\label{tab:domains}
\end{table*}

%% file: Tables/prompt_eval_groundedness.tex
\begin{table*}[htbp]
\scriptsize
\centering
\begin{tabular}{@{}p{\linewidth}@{}}
\toprule
\textbf{Supportedness Evaluation Instruction}\\
\midrule
Instruction:\\
1. You will be given a question, a statement, and an external document.\\
2. First, extract all subclaims within the statement that need verification.\\
3. Assess how well each subclaim is supported by the document.\\
4. Assign one of the following labels: "fully support," "partially support," or "not support."\\
   - If all subclaims are supported by the document, select "fully support."\\
   - If only some of the subclaims are supported, select "partially support."\\
   - If none of the subclaims are supported, select "not support."\\
\\
Important:\\
Provide a brief explanation for your chosen level of support. The final answer should begin with "Answer: ".\\
\\
Statement: \{statement\}\\
Documents: \{document\}\\
\bottomrule
\end{tabular}
    \caption{Prompt template for supportedness evaluator.}
    \label{tab:groundedness}
\end{table*}

%% file: Tables/prompt_eval_completeness.tex
\begin{table*}[htbp]
\scriptsize
\centering
\begin{tabular}{@{}p{\linewidth}@{}}
\toprule
\textbf{Completeness Evaluation Instruction}\\
\midrule
Instruction:\\
1. You will be given a fact and model response.\\
2. Evaluate how thoroughly the fact is addressed by the model response.\\
3. Assign one of the following labels:\\
\quad - Fully addressed: The fact is completely addressed in the model response. The details of the fact are fully supported by the model response.\\
\quad - Partially addressed: The fact is addressed to some extent, but important details are missing or insufficiently supported. Some details of the fact are not supported by the model response.\\
\quad - Not addressed: The fact is not addressed at all in the model response.\\
\\
Important: The final answer should begin with 'Label:', and must not include any other text.\\
\\
Fact: \{fact\}\\
Statement: \{statement\}\\
\bottomrule
\end{tabular}
    \caption{Prompt template for completeness evaluator.}
    \label{tab:completeness}
\end{table*}

%% file: Tables/prompt_eval_relevance.tex
\begin{table*}[htbp]
\scriptsize
\centering
\begin{tabular}{@{}p{\linewidth}@{}}
\toprule
\textbf{Relevance Evaluation Instruction}\\
\midrule
Instructions:\\
1. You will be given a question and a statement.\\
2. Evaluate how the statement is related to the question.\\
3. Assign one of the following labels to each subclaim:\\
\quad - Fully relevant: The statement directly addresses the question.\\
\quad - Partially relevant: The statement is somewhat related to the question.\\
\quad - Not relevant: The statement is unrelated to the question.\\
\\
Important:\\
Provide a brief explanation for your chosen level of relevance. The final label should begin with 'Label:'.\\
\\
Question: $<$image$>$\{question\}\\
Statement: \{statement\}\\
\bottomrule
\end{tabular}
    \caption{Prompt template for relevance evaluator.}
    \label{tab:relevance}
\end{table*}

%% file: Tables/prompt_query_generation.tex
\begin{table*}[htbp]
\scriptsize
\centering
\begin{tabular}{@{}p{\linewidth}@{}}
\toprule
\textbf{Query Generation Instruction}\\
\midrule
$<$Instruction$>$\\
1. Based on the given image and question, generate \{N\} search queries.\\
2. Formulate queries to retrieve documents that provide information to generate the answer.\\
3. List the generated search queries separated by commas. For example: ``query 1'', ``query 2'', ...\\
\\
Question: $<$image$>$\{question\}\\
Search queries: \\
\bottomrule
\end{tabular}
    \caption{Instruction for search query generation.}
    \label{tab:prompt_query}
\end{table*}

%% file: Tables/prompt_vanilla.tex
\begin{table*}[htbp]
\scriptsize
\centering
\begin{tabular}{@{}p{\linewidth}@{}}
\toprule
\textbf{Vanilla Instruction}\\
\midrule
Based on the documents, provide a helpful answer in paragraph form. Your answer must be supported by the content in the citations. \\
You should cite the passage number (indices) in the format of [1][2][3] at the end of each sentence. \\
Do not include sentences that are not supported by the documents.
\\
\\
Question: $<$image$>$\{question\}\ \\
Document: \\
...\\
\\
Answer: \\
\bottomrule
\end{tabular}
    \caption{Instruction for \textsc{Vanilla} method.}
    \label{tab:prompt_answer}
\end{table*}

%% file: Tables/prompt_cot.tex
\begin{table*}[htbp]
\scriptsize
\centering
\begin{tabular}{@{}p{\linewidth}@{}}
\toprule
\textbf{CoT Instruction}\\
\midrule
Your task is to answer the question using the provided documents and cite your answer with their passage numbers.\\
\\
First, before answering, show your reasoning process using the $<$thinking$>$ and $</$thinking$>$ tags. Follow the thinking process format below: \\
\qquad 1. Identify the relevant documents related to the question: "The only relevant documents to the question are documents [$<$relevant\_doc$>$], [$<$relevant\_doc$>$].." \\
\qquad 2. Analyze the relevant information from the identified documents: "From document [$<$relevant\_doc$>$], we know that '$<$relevant\_info$>$', '$<$relevant\_info$>$'.." \\
\\
Second, answer the question, explicitly incorporating copied spans into your answer. Your answer must be supported by the content in the citations. You should cite the passage number (indices) in the format of [1][2][3] at the end of the sentence. Do not refer to the documents explicitly using phrases like "Document [5] states" or "According to Document [3]". Here is an output example:\\
--\\
$<$thinking$>$\\
1. Identify the relevant documents related to the question: "The only relevant documents to the question are documents [1], [2], [3], [5]."\\
2. Analyze the relevant information from the identified documents: \\
\qquad- From document [1], we know that "traditional, low-sugar and no-sugar fruit preserving methods including bottling, jams and jellies, fruit pie fillings, dehydrating and cooking" can be used.\\
\qquad- From document [2], we learn about "fresh fruit desserts, jams, jellies, preserves, canned peaches, pears, cherries and apricots, and fresh fruit salads," and tips for "canning, freezing fruit and keeping fruit fresh."\\
\qquad- From document [3], grilling is suggested as a creative method: "Grilling stone fruits only serves to heighten their natural sweetness."\\
\qquad- From document [5], apricots can be "dried, cooked into pastry, and eaten as jam" or "distilled into brandy and liqueur."\\
$<$/thinking$>$\\
--\\
\\
Question: $<$image$>$\{question\}\ \\
Document: \\
...\\
\\
Let's think step-by-step: \\
\bottomrule
\end{tabular}
\caption{Instruction for \textsc{Chain-of-Thought} method.}
\label{tab:prompt_cot}
\end{table*}

%% file: Tables/prompt_agg.tex
\begin{table*}[htbp]
\scriptsize
\centering
\begin{tabular}{@{}p{\linewidth}@{}}
\toprule
\textbf{Knowledge Extraction Instruction}\\
\midrule
Your task is to extract factual information from the provided document. Include only details that can be confidently determined, excluding imaginary, speculative, or aesthetic content. Present the information clearly and concisely in paragraph form. Do not explicitly refer to the document itself or use introductory phrases such as "the document states," "it mentions," or "according to the document." Instead, directly state the factual information.\\
\\
Document:\\
\{document\}\\
\bottomrule
\end{tabular}
\caption{Instruction for \textsc{Knowledge Extraction} method.}
\label{tab:prompt_agg}
\end{table*}

%% file: aaai2026.bib
@article{chen2024we,
  title={Are we on the right way for evaluating large vision-language models?},
  author={Chen, Lin and Li, Jinsong and Dong, Xiaoyi and Zhang, Pan and Zang, Yuhang and Chen, Zehui and Duan, Haodong and Wang, Jiaqi and Qiao, Yu and Lin, Dahua and others},
  journal={arXiv preprint arXiv:2403.20330},
  year={2024}
}

@article{hurst2024gpt,
  title={Gpt-4o system card},
  author={Hurst, Aaron and Lerer, Adam and Goucher, Adam P and Perelman, Adam and Ramesh, Aditya and Clark, Aidan and Ostrow, AJ and Welihinda, Akila and Hayes, Alan and Radford, Alec and others},
  journal={arXiv preprint arXiv:2410.21276},
  year={2024}
}

@article{min2023factscore,
  title={Factscore: Fine-grained atomic evaluation of factual precision in long form text generation},
  author={Min, Sewon and Krishna, Kalpesh and Lyu, Xinxi and Lewis, Mike and Yih, Wen-tau and Koh, Pang Wei and Iyyer, Mohit and Zettlemoyer, Luke and Hajishirzi, Hannaneh},
  journal={arXiv preprint arXiv:2305.14251},
  year={2023}
}

@article{jiang2024mixtral,
  title={Mixtral of experts},
  author={Jiang, Albert Q and Sablayrolles, Alexandre and Roux, Antoine and Mensch, Arthur and Savary, Blanche and Bamford, Chris and Chaplot, Devendra Singh and Casas, Diego de las and Hanna, Emma Bou and Bressand, Florian and others},
  journal={arXiv preprint arXiv:2401.04088},
  year={2024}
}

@article{abdin2024phi,
  title={Phi-4 technical report},
  author={Abdin, Marah and Aneja, Jyoti and Behl, Harkirat and Bubeck, S{\'e}bastien and Eldan, Ronen and Gunasekar, Suriya and Harrison, Michael and Hewett, Russell J and Javaheripi, Mojan and Kauffmann, Piero and others},
  journal={arXiv preprint arXiv:2412.08905},
  year={2024}
}

@inproceedings{chen2024internvl,
  title={Internvl: Scaling up vision foundation models and aligning for generic visual-linguistic tasks},
  author={Chen, Zhe and Wu, Jiannan and Wang, Wenhai and Su, Weijie and Chen, Guo and Xing, Sen and Zhong, Muyan and Zhang, Qinglong and Zhu, Xizhou and Lu, Lewei and others},
  booktitle={Proceedings of the IEEE/CVF conference on computer vision and pattern recognition},
  pages={24185--24198},
  year={2024}
}

@inproceedings{chang2022webqa,
  title={Webqa: Multihop and multimodal qa},
  author={Chang, Yingshan and Narang, Mridu and Suzuki, Hisami and Cao, Guihong and Gao, Jianfeng and Bisk, Yonatan},
  booktitle={Proceedings of the IEEE/CVF conference on computer vision and pattern recognition},
  pages={16495--16504},
  year={2022}
}

@article{dubey2024llama,
  title={The llama 3 herd of models},
  author={Dubey, Abhimanyu and Jauhri, Abhinav and Pandey, Abhinav and Kadian, Abhishek and Al-Dahle, Ahmad and Letman, Aiesha and Mathur, Akhil and Schelten, Alan and Yang, Amy and Fan, Angela and others},
  journal={arXiv e-prints},
  pages={arXiv--2407},
  year={2024}
}

@article{kulkarni2020aquamuse,
  title={Aquamuse: Automatically generating datasets for query-based multi-document summarization},
  author={Kulkarni, Sayali and Chammas, Sheide and Zhu, Wan and Sha, Fei and Ie, Eugene},
  journal={arXiv preprint arXiv:2010.12694},
  year={2020}
}

@article{fan2019eli5,
  title={ELI5: Long form question answering},
  author={Fan, Angela and Jernite, Yacine and Perez, Ethan and Grangier, David and Weston, Jason and Auli, Michael},
  journal={arXiv preprint arXiv:1907.09190},
  year={2019}
}

@article{boni2021howsumm,
  title={Howsumm: A multi-document summarization dataset derived from wikihow articles},
  author={Boni, Odellia and Feigenblat, Guy and Lev, Guy and Shmueli-Scheuer, Michal and Sznajder, Benjamin and Konopnicki, David},
  journal={arXiv preprint arXiv:2110.03179},
  year={2021}
}

@inproceedings{bolotova2023wikihowqa,
  title={WikiHowQA: A comprehensive benchmark for multi-document non-factoid question answering},
  author={Bolotova-Baranova, Valeriia and Blinov, Vladislav and Filippova, Sofya and Scholer, Falk and Sanderson, Mark},
  booktitle={Proceedings of the 61st Annual Meeting of the Association for Computational Linguistics (Volume 1: Long Papers)},
  pages={5291--5314},
  year={2023}
}

@article{han2024rag,
  title={Rag-qa arena: Evaluating domain robustness for long-form retrieval augmented question answering},
  author={Han, Rujun and Zhang, Yuhao and Qi, Peng and Xu, Yumo and Wang, Jenyuan and Liu, Lan and Wang, William Yang and Min, Bonan and Castelli, Vittorio},
  journal={arXiv preprint arXiv:2407.13998},
  year={2024}
}

@article{wei2024long,
  title={Long-form factuality in large language models},
  author={Wei, Jerry and Yang, Chengrun and Song, Xinying and Lu, Yifeng and Hu, Nathan and Huang, Jie and Tran, Dustin and Peng, Daiyi and Liu, Ruibo and Huang, Da and others},
  journal={arXiv preprint arXiv:2403.18802},
  year={2024}
}

@article{huh2024long,
  title={Long-Form Answers to Visual Questions from Blind and Low Vision People},
  author={Huh, Mina and Xu, Fangyuan and Peng, Yi-Hao and Chen, Chongyan and Murugu, Hansika and Gurari, Danna and Choi, Eunsol and Pavel, Amy},
  journal={arXiv preprint arXiv:2408.06303},
  year={2024}
}

@inproceedings{lin2004rouge,
  title={Rouge: A package for automatic evaluation of summaries},
  author={Lin, Chin-Yew},
  booktitle={Text summarization branches out},
  pages={74--81},
  year={2004}
}

@article{jing2023faithscore,
  title={FaithScore: Fine-grained Evaluations of Hallucinations in Large Vision-Language Models},
  author={Jing, Liqiang and Li, Ruosen and Chen, Yunmo and Du, Xinya},
  journal={arXiv preprint arXiv:2311.01477},
  year={2023}
}

@article{raffel2020exploring,
  title={Exploring the limits of transfer learning with a unified text-to-text transformer},
  author={Raffel, Colin and Shazeer, Noam and Roberts, Adam and Lee, Katherine and Narang, Sharan and Matena, Michael and Zhou, Yanqi and Li, Wei and Liu, Peter J},
  journal={Journal of machine learning research},
  volume={21},
  number={140},
  pages={1--67},
  year={2020}
}

@inproceedings{yue2024mmmu,
  title={Mmmu: A massive multi-discipline multimodal understanding and reasoning benchmark for expert agi},
  author={Yue, Xiang and Ni, Yuansheng and Zhang, Kai and Zheng, Tianyu and Liu, Ruoqi and Zhang, Ge and Stevens, Samuel and Jiang, Dongfu and Ren, Weiming and Sun, Yuxuan and others},
  booktitle={Proceedings of the IEEE/CVF Conference on Computer Vision and Pattern Recognition},
  pages={9556--9567},
  year={2024}
}

@article{zhou2023marvel,
  title={MARVEL: unlocking the multi-modal capability of dense retrieval via visual module plugin},
  author={Zhou, Tianshuo and Mei, Sen and Li, Xinze and Liu, Zhenghao and Xiong, Chenyan and Liu, Zhiyuan and Gu, Yu and Yu, Ge},
  journal={arXiv preprint arXiv:2310.14037},
  year={2023}
}

@inproceedings{papineni2002bleu,
  title={Bleu: a method for automatic evaluation of machine translation},
  author={Papineni, Kishore and Roukos, Salim and Ward, Todd and Zhu, Wei-Jing},
  booktitle={Proceedings of the 40th annual meeting of the Association for Computational Linguistics},
  pages={311--318},
  year={2002}
}

@article{liu2022universal,
  title={Universal vision-language dense retrieval: Learning a unified representation space for multi-modal retrieval},
  author={Liu, Zhenghao and Xiong, Chenyan and Lv, Yuanhuiyi and Liu, Zhiyuan and Yu, Ge},
  journal={arXiv preprint arXiv:2209.00179},
  year={2022}
}

@article{lin2024mm,
  title={Mm-embed: Universal multimodal retrieval with multimodal llms},
  author={Lin, Sheng-Chieh and Lee, Chankyu and Shoeybi, Mohammad and Lin, Jimmy and Catanzaro, Bryan and Ping, Wei},
  journal={arXiv preprint arXiv:2411.02571},
  year={2024}
}

@article{overwijk2022clueweb22,
  title={Clueweb22: 10 billion web documents with visual and semantic information},
  author={Overwijk, Arnold and Xiong, Chenyan and Liu, Xiao and VandenBerg, Cameron and Callan, Jamie},
  journal={arXiv preprint arXiv:2211.15848},
  year={2022}
}

@inproceedings{wei2024uniir,
  title={Uniir: Training and benchmarking universal multimodal information retrievers},
  author={Wei, Cong and Chen, Yang and Chen, Haonan and Hu, Hexiang and Zhang, Ge and Fu, Jie and Ritter, Alan and Chen, Wenhu},
  booktitle={European Conference on Computer Vision},
  pages={387--404},
  year={2024},
  organization={Springer}
}

@article{2025skyworkvlrm,
  title={Skywork-vl reward: An effective reward model for multimodal understanding and reasoning},
  author={Wang, Xiaokun and Wang, Peiyu and Pei, Jiangbo and Shen, Wei and Peng, Yi and Hao, Yunzhuo and Qiu, Weijie and Jian, Ai and Xie, Tianyidan and Song, Xuchen and others},
  journal={arXiv preprint arXiv:2505.07263},
  year={2025}
}

@article{he2021debertav3,
  title={Debertav3: Improving deberta using electra-style pre-training with gradient-disentangled embedding sharing},
  author={He, Pengcheng and Gao, Jianfeng and Chen, Weizhu},
  journal={arXiv preprint arXiv:2111.09543},
  year={2021}
}

@article{sanyal2024machines,
  title={Are Machines Better at Complex Reasoning? Unveiling Human-Machine Inference Gaps in Entailment Verification},
  author={Sanyal, Soumya and Xiao, Tianyi and Liu, Jiacheng and Wang, Wenya and Ren, Xiang},
  journal={arXiv preprint arXiv:2402.03686},
  year={2024}
}

@inproceedings{wang2022ofa,
  title={Ofa: Unifying architectures, tasks, and modalities through a simple sequence-to-sequence learning framework},
  author={Wang, Peng and Yang, An and Men, Rui and Lin, Junyang and Bai, Shuai and Li, Zhikang and Ma, Jianxin and Zhou, Chang and Zhou, Jingren and Yang, Hongxia},
  booktitle={International conference on machine learning},
  pages={23318--23340},
  year={2022},
  organization={PMLR}
}

@article{wang2024qwen2,
  title={Qwen2-vl: Enhancing vision-language model's perception of the world at any resolution},
  author={Wang, Peng and Bai, Shuai and Tan, Sinan and Wang, Shijie and Fan, Zhihao and Bai, Jinze and Chen, Keqin and Liu, Xuejing and Wang, Jialin and Ge, Wenbin and others},
  journal={arXiv preprint arXiv:2409.12191},
  year={2024}
}

@article{yang2025qwen3,
  title={Qwen3 technical report},
  author={Yang, An and Li, Anfeng and Yang, Baosong and Zhang, Beichen and Hui, Binyuan and Zheng, Bo and Yu, Bowen and Gao, Chang and Huang, Chengen and Lv, Chenxu and others},
  journal={arXiv preprint arXiv:2505.09388},
  year={2025}
}

@article{li2023self,
  title={Self-checker: Plug-and-play modules for fact-checking with large language models},
  author={Li, Miaoran and Peng, Baolin and Galley, Michel and Gao, Jianfeng and Zhang, Zhu},
  journal={arXiv preprint arXiv:2305.14623},
  year={2023}
}

@article{pillutla2021mauve,
  title={Mauve: Measuring the gap between neural text and human text using divergence frontiers},
  author={Pillutla, Krishna and Swayamdipta, Swabha and Zellers, Rowan and Thickstun, John and Welleck, Sean and Choi, Yejin and Harchaoui, Zaid},
  journal={Advances in Neural Information Processing Systems},
  volume={34},
  pages={4816--4828},
  year={2021}
}

@article{gao2023enabling,
  title={Enabling large language models to generate text with citations},
  author={Gao, Tianyu and Yen, Howard and Yu, Jiatong and Chen, Danqi},
  journal={arXiv preprint arXiv:2305.14627},
  year={2023}
}

@article{bohnet2022attributed,
  title={Attributed question answering: Evaluation and modeling for attributed large language models},
  author={Bohnet, Bernd and Tran, Vinh Q and Verga, Pat and Aharoni, Roee and Andor, Daniel and Soares, Livio Baldini and Ciaramita, Massimiliano and Eisenstein, Jacob and Ganchev, Kuzman and Herzig, Jonathan and others},
  journal={arXiv preprint arXiv:2212.08037},
  year={2022}
}

@article{slobodkin2024attribute,
  title={Attribute first, then generate: Locally-attributable grounded text generation},
  author={Slobodkin, Aviv and Hirsch, Eran and Cattan, Arie and Schuster, Tal and Dagan, Ido},
  journal={arXiv preprint arXiv:2403.17104},
  year={2024}
}

@article{li2024improving,
  title={Improving attributed text generation of large language models via preference learning},
  author={Li, Dongfang and Sun, Zetian and Hu, Baotian and Liu, Zhenyu and Hu, Xinshuo and Liu, Xuebo and Zhang, Min},
  journal={arXiv preprint arXiv:2403.18381},
  year={2024}
}

@article{sun2023towards,
  title={Towards verifiable text generation with evolving memory and self-reflection},
  author={Sun, Hao and Cai, Hengyi and Wang, Bo and Hou, Yingyan and Wei, Xiaochi and Wang, Shuaiqiang and Zhang, Yan and Yin, Dawei},
  journal={arXiv preprint arXiv:2312.09075},
  year={2023}
}

@article{li2024think,
  title={Think\&Cite: Improving Attributed Text Generation with Self-Guided Tree Search and Progress Reward Modeling},
  author={Li, Junyi and Ng, Hwee Tou},
  journal={arXiv preprint arXiv:2412.14860},
  year={2024}
}

@article{huang2024training,
  title={Training language models to generate text with citations via fine-grained rewards},
  author={Huang, Chengyu and Wu, Zeqiu and Hu, Yushi and Wang, Wenya},
  journal={arXiv preprint arXiv:2402.04315},
  year={2024}
}

@article{ye2023cognitive,
  title={Cognitive mirage: A review of hallucinations in large language models},
  author={Ye, Hongbin and Liu, Tong and Zhang, Aijia and Hua, Wei and Jia, Weiqiang},
  journal={arXiv preprint arXiv:2309.06794},
  year={2023}
}

@article{zhang2023siren,
  title={Siren's song in the AI ocean: a survey on hallucination in large language models},
  author={Zhang, Yue and Li, Yafu and Cui, Leyang and Cai, Deng and Liu, Lemao and Fu, Tingchen and Huang, Xinting and Zhao, Enbo and Zhang, Yu and Chen, Yulong and others},
  journal={arXiv preprint arXiv:2309.01219},
  year={2023}
}

@article{huang2025survey,
  title={A survey on hallucination in large language models: Principles, taxonomy, challenges, and open questions},
  author={Huang, Lei and Yu, Weijiang and Ma, Weitao and Zhong, Weihong and Feng, Zhangyin and Wang, Haotian and Chen, Qianglong and Peng, Weihua and Feng, Xiaocheng and Qin, Bing and others},
  journal={ACM Transactions on Information Systems},
  volume={43},
  number={2},
  pages={1--55},
  year={2025},
  publisher={ACM New York, NY}
}

@article{liu2023evaluating,
  title={Evaluating verifiability in generative search engines},
  author={Liu, Nelson F and Zhang, Tianyi and Liang, Percy},
  journal={arXiv preprint arXiv:2304.09848},
  year={2023}
}

@article{li2024llava,
  title={Llava-onevision: Easy visual task transfer},
  author={Li, Bo and Zhang, Yuanhan and Guo, Dong and Zhang, Renrui and Li, Feng and Zhang, Hao and Zhang, Kaichen and Zhang, Peiyuan and Li, Yanwei and Liu, Ziwei and others},
  journal={arXiv preprint arXiv:2408.03326},
  year={2024}
}

@article{berchansky2024cotar,
  title={CoTAR: Chain-of-Thought Attribution Reasoning with Multi-level Granularity},
  author={Berchansky, Moshe and Fleischer, Daniel and Wasserblat, Moshe and Izsak, Peter},
  journal={arXiv preprint arXiv:2404.10513},
  year={2024}
}

@article{wei2022chain,
  title={Chain-of-thought prompting elicits reasoning in large language models},
  author={Wei, Jason and Wang, Xuezhi and Schuurmans, Dale and Bosma, Maarten and Xia, Fei and Chi, Ed and Le, Quoc V and Zhou, Denny and others},
  journal={Advances in neural information processing systems},
  volume={35},
  pages={24824--24837},
  year={2022}
}

@inproceedings{deng2025words,
  title={Words or Vision: Do Vision-Language Models Have Blind Faith in Text?},
  author={Deng, Ailin and Cao, Tri and Chen, Zhirui and Hooi, Bryan},
  booktitle={Proceedings of the Computer Vision and Pattern Recognition Conference},
  pages={3867--3876},
  year={2025}
}

@inproceedings{yu2025mramg,
  title={MRAMG-Bench: A Comprehensive Benchmark for Advancing Multimodal Retrieval-Augmented Multimodal Generation},
  author={Yu, Qinhan and Xiao, Zhiyou and Li, Binghui and Wang, Zhengren and Chen, Chong and Zhang, Wentao},
  booktitle={Proceedings of the 48th International ACM SIGIR Conference on Research and Development in Information Retrieval},
  pages={3616--3626},
  year={2025}
}

@article{ma2024multi,
  title={Multi-modal Retrieval Augmented Multi-modal Generation: Datasets, Evaluation Metrics and Strong Baselines},
  author={Ma, Zi-Ao and Lan, Tian and Tu, Rong-Cheng and Hu, Yong and Zhu, Yu-Shi and Zhang, Tong and Huang, Heyan and Wu, Zhijing and Mao, Xian-Ling},
  journal={arXiv preprint arXiv:2411.16365},
  year={2024}
}

@article{wu2025language,
  title={When Language Overrules: Revealing Text Dominance in Multimodal Large Language Models},
  author={Wu, Huyu and Tang, Meng and Zheng, Xinhan and Jiang, Haiyun},
  journal={arXiv preprint arXiv:2508.10552},
  year={2025}
}

@article{loper2002nltk,
  title={Nltk: The natural language toolkit},
  author={Loper, Edward and Bird, Steven},
  journal={arXiv preprint cs/0205028},
  year={2002}
}

@article{wolf2019huggingface,
  title={Huggingface's transformers: State-of-the-art natural language processing},
  author={Wolf, Thomas and Debut, Lysandre and Sanh, Victor and Chaumond, Julien and Delangue, Clement and Moi, Anthony and Cistac, Pierric and Rault, Tim and Louf, R{\'e}mi and Funtowicz, Morgan and others},
  journal={arXiv preprint arXiv:1910.03771},
  year={2019}
}
